\documentclass{article} 
\usepackage{iclr2025_conference,times}
\usepackage{makecell}
\usepackage{fourier} 
\usepackage{xcolor}
\usepackage{colortbl}
\usepackage{multirow}
\usepackage{nicematrix}

\usepackage{amsmath,amsfonts,bm}

\def\eqref#1{equation~\ref{#1}}

\def\1{\bm{1}}

\DeclareMathAlphabet{\mathsfit}{\encodingdefault}{\sfdefault}{m}{sl}
\SetMathAlphabet{\mathsfit}{bold}{\encodingdefault}{\sfdefault}{bx}{n}

\usepackage{hyperref}
\usepackage{url}

\title{Foundation Models and Transformers for Anomaly Detection: A Survey}

\author{Mou\"{i}n Ben~Ammar\textsuperscript{$\diamond,\dagger$,$*$},  Arturo Mendoza\textsuperscript{$\dagger$}, Nacim Belkhir\textsuperscript{$\S$}, Antoine Manzanera\textsuperscript{$\diamond$},  Gianni Franchi\textsuperscript{$\diamond$}
\thanks{corresponding author} \\
U2IS Lab ENSTA Paris\textsuperscript{$\diamond$},  Palaiseau, FRANCE \\
SafranTech\textsuperscript{$\dagger$}, Chateaufort 78117, FRANCE \\
INRIA Saclay \textsuperscript{$\S$}, Palaiseau, France \\
\texttt{\{first.last\}@enstaparis.fr\textsuperscript{$\diamond$}, safrangroup.com\textsuperscript{$\dagger$}, inria.fr\textsuperscript{$\S$} }
}

\newcommand{\keywords}[1]{%
  \begin{center}
  \textbf{Keywords:} #1
  \end{center}
}

\newcommand{\ie}{\textit{i.e.}, }
\iclrfinalcopy 
\begin{document}

\maketitle

\begin{abstract}

In line with the development of deep learning, this survey examines the transformative role of Transformers and foundation models in advancing visual anomaly detection (VAD). We explore how these architectures, with their global receptive fields and adaptability, address challenges such as long-range dependency modeling, contextual modeling and data scarcity. The survey categorizes VAD methods into reconstruction-based, feature-based and zero/few-shot approaches, highlighting the paradigm shift brought about by foundation models. By integrating attention mechanisms and leveraging large-scale pre-training, Transformers and foundation models enable more robust, interpretable, and scalable anomaly detection solutions. This work provides a comprehensive review of state-of-the-art techniques, their strengths, limitations, and emerging trends in leveraging these architectures for VAD.

\end{abstract}
\keywords{
Anomaly detection Transformers Foundation models deep learning computer vision unsupervised learning self-supervised learning  Survey
}

A common requirement when analyzing real world data is to identify deviations from the expected patterns of the data. These deviations are referred to as anomalies, and usually constitutes an important event that requires attention, due to the significant repercussion it may entail. In recent years, anomaly detection (AD) has become increasingly popular, with applications spanning various domains including finance~\cite{AHMED2016278}, cybersecurity~\cite{AHMED201619}, insurance~\cite{HILAL2022116429}, healthcare~\cite{10.1145/3464423}, manufacturing~\cite{Liu_2024} and surveillance~\cite{ANOOPA2022162}.
When applied on vision systems, \emph{visual anomaly detection (VAD)} aims to detect these deviations and localize them.

The rise of deep learning \cite{8187120,lecun2015deep}, in particular with the introduction of Convolution Neural Networks (CNNs) \cite{5537907} allowed for rapid improvement in the VAD field, with a plethora of algorithmic solutions to address the diverse challenges that may arise.
However, despite the considerable contributions CNNs have provided, these architectures intrinsically lack some properties that are required for a robust VAD. Mainly, the inherent locality of the convolution operation due to their limited receptive field \cite{xu2021limits}, which impedes modeling long-range relations. This weakness becomes most apparent with global and logical anomalies, and is further aggravated by their spatial invariance \cite{alsallakh2020mind}.
Multiple works mitigated these issues~\cite{Roth_2022_CVPR,Batzner_2024_WACV,Bae_2023_ICCV,Zavrtanik_2021_ICCV,defard2021padim}, by designing algorithms to counteract the architecture intrinsic biases.

The introduction of Transformers into the vision field by \citet{dosovitskiy2021imageworth16x16words} provided more direct solutions to the previously cited issues, as they proved more suitable to these tasks. 

Transformers are equipped with a global receptive field thanks to the attention mechanism. This provides richer global and local contextual information to model complex spatial relationships and capture fine-grained details in the input image.
Additionally, due to their unique structure and ease of adaptation through prompt tuning~\cite{Sohn_2023_CVPR,10.1007/978-3-031-19827-4_41,chen2024humanfreeautomatedpromptingvisionlanguage,Li_2024_CVPR,pmlr-v162-he22f,pmlr-v202-yoo23a,liu2021p}, they are more adapted to tackle challenging tasks such as multi-class and multimodal AD, with an additional layer of interpretability.

Transformers are more capable of extracting useful information for downstream tasks through large-scale pre-training. This quality has contributed to the progressive advancement towards the concept of \emph{foundation models}, often built using Transformer architectures pre-trained on Web-scale datasets. These models offer an unprecedented versatility and efficiency in solving multiple downstream-tasks, including VAD.
The introduction of foundation models have contributed in a paradigm shift in the field of VAD. Traditionally, VAD algorithms are conceptualized and classified through the degree of supervision they are provided for training. However, with foundation models, the task has shifted towards zero-shot anomaly detection (ZSAD). Leveraging these models relaxes the data and modeling constraints of the VAD task while maintaining strong performance.

Due to their strength and versatility, many strategies have been proposed to leverage Transformers and foundation models in order to tackle the many challenges within VAD.
Prior surveys have covered various aspects of AD, including unsupervised methods~\cite{cui2023survey}, self-supervised methods~\cite{hojjati2024self}, generative models~\cite{sabuhi2021applications,xia2022gan} and many other aspects of anomaly detection \cite{su2024large,liu2024deep,liu2024generalized,liu2025networking,liu2024privacy,liu2025survey}.
For example, \citet{liu2025survey} covers the use of diffusion models, while \citet{liu2024privacy} focuses on the privacy preservation aspect, and \cite{liu2025networking} focuses on AD for Networking Systems on video surveillance.

A survey tackling the paradigm shift with the introduction of Transformer-based models, and their extension to foundation models is still lacking. Concretely, on how the architectural and learning differences between Transformers and CNN models considerably affect the VAD algorithms.

Although the newly introduced survey by \citet{miyai2024generalizedoutofdistributiondetectionvision} addresses anomaly detection in the Large Vision-Language Models (VLMs) era, particularly CLIP. The authors cover solely works on image data, focusing mostly on OOD detection and not including general Transformer models. 
This survey aims to bridge this gap by providing a structured taxonomy of a wider range of visual-AD, including video and point cloud methods leveraging attention-based architectures. Our contribution lies in systematically reviewing reconstruction-based, feature-based, zero- and few-shot detection methods, discussing the unique challenges, strengths, and future directions of these models. 

%

The aim of this survey is two-fold, firstly to present a structured and comprehensive review of research methods in VAD based on Transformers and attention models, highlighting the methodological changes due to this transition from convolutional models. Secondly, to explore the recent advances in VAD methods leveraging foundation models and their conceptual design.
Delving into the fundamental concepts behind these models, and the specific details of their integration into the VAD field. Then exploring various approaches depending on the design of the detection algorithm and the adopted architecture, and discuss their strengths and limitations.

The remainder of this survey will be organized as follows: First, we explain the Transformer architecture in section \ref{subsec:transformer}, its uniqueness and the many variants of such architecture. Section \ref{subsec:foundation} defines the concepts behind 
foundation models, and provides a comprehensive list of their wide applicability range. 
In section~\ref{section:Defs}, we provide precise definitions for the various disciplines aiming to detect unknowns, their similarities and distinctions compared to anomaly detection.
Section~\ref{sec:taxonomy} outlines our survey and details the adopted taxonomy for AD methods

We focus on the case where labels are unavailable, \ie the unsupervised and self-supervised learning cases, which is more aligned with real-word scenarios and we divide it into many sub-categories:  Reconstruction based methods are detailed in section~\ref{sec:reconstuction_based},  with Feature-based methods presented in section~\ref{sec:features}) and the Zero- and Few-shot shot cases are tackled in
section~\ref{sec:zero_few_shot}.

Figure~\ref{fig:taxonomy} illustrates the adopted taxonomy in this survey, and Table \ref{tab:anomaly-detection} provides the list of methods presented at each sub-category.

\section{Transformers and Foundation models}

\subsection{Transformer Architectures}
\label{subsec:transformer}
Transformer models have brought about a significant shift in the landscape of natural language processing (NLP), completely transforming tasks like machine translation, text generation, and sentiment analysis.

\begin{figure}[!ht]
    \centering
    \includegraphics[width=0.6\linewidth]{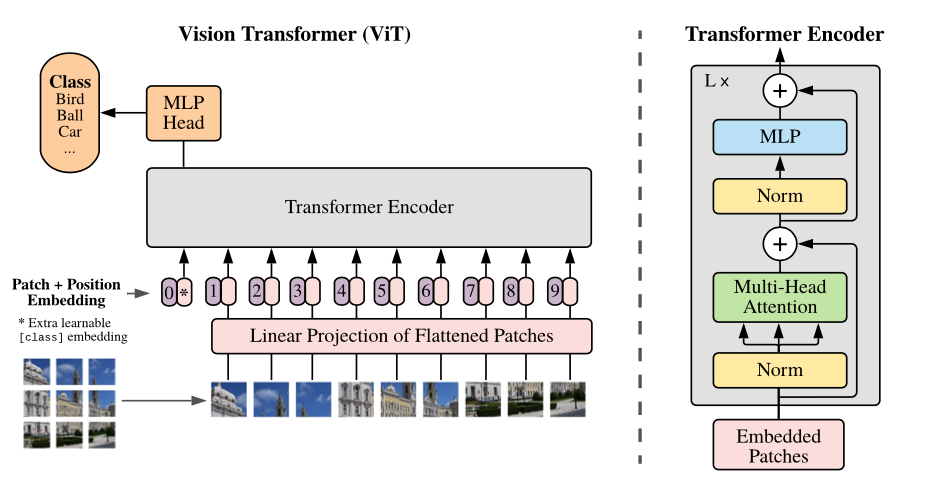}
    \includegraphics[width=0.35\linewidth]{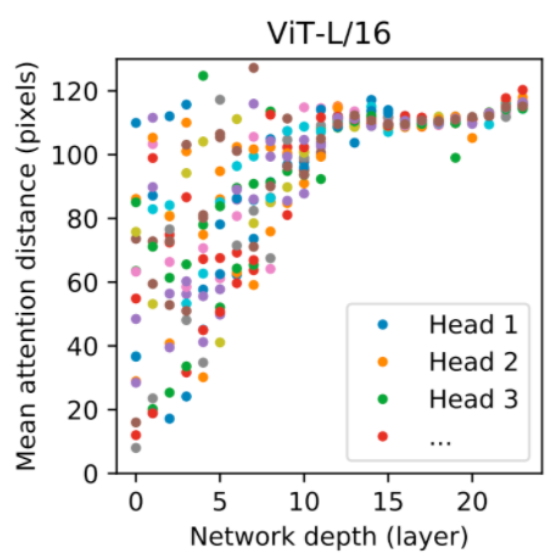}
    \caption{Vision Transformer ViT Architecture (left). Mean size of attended area (receptive field) by head and network depth (right)~\cite{dosovitskiy2021image}.}%
    \label{fig:vit-figure}
\end{figure}

Since their inception by \citet{vaswani2023attention} in 2017, Transformers have emerged as a cornerstone in NLP, outperforming traditional approaches like recurrent neural networks and convolutional neural networks (CNNs) by effectively capturing long-range dependencies in sequential data. This paved the way for the development of various Transformer-based models such as BERT~\cite{devlin2019bert}, GPT variants~\cite{openai2024gpt4,brown2020language,Radford2019LanguageMA,radford2018improving},  T5 (Text-to-Text Transfer Transformer)~\cite{2020t5}, BLOOM~\cite{bigscience_workshop_2022}, among others. For a deeper dive into these models, we recommend referring to surveys that provide comprehensive insights into Transformers for NLP~\cite{Tay2020EfficientTA,hu2018introductory,kalyan2021ammus}. 
Our focus however lies on their application in vision tasks.
The remarkable achievements of Transformers in NLP have sparked interest within the computer vision community to adapt these models for vision and multimodal learning tasks. Nevertheless, visual data follows its own structured patterns, such as spatial and temporal coherence, necessitating novel network architectures and learning strategies. One common technique involves replacing words in NLP sequences with image patches, as exemplified by the Vision Transformer (ViT) approach~\cite{dosovitskiy2021image} (see figure \ref{fig:vit-figure}).


\subsubsection{Layers in Transformers}
\label{Layers_in_Transformers}

In the field of vision, Transformer architectures primarily rely on three key 
mechanisms:
Self attention layer, Masked attention layer, and Cross attention layer.

The \textbf{Self-Attention layer} plays a pivotal role in Transformers within the domain of computer vision. It evaluates the relevance of each token concerning other elements within a sequence, elucidating which components are likely to be correlated. This mechanism is fundamental for structured prediction tasks, as it explicitly models interactions among all entities in a sequence. Let's denote the sequence as ($\mathbf{x}_1, \mathbf{x}_2, \cdots \mathbf{x}_n$), represented by $\mathbf{X} \in \mathbb{R}^{n \times d}$, where $d$ 
is
the dimension used to represent each entity.
To operate, the Self-Attention layer necessitates three types of inputs: input queries $Q$ and key-value pairs $(K, V)$, which are constructed using three matrices ($\mathbf{W}^V \in \mathbb{R}^{d \times d_v}$), ($\mathbf{W}^Q \in \mathbb{R}^{d \times d_q}$), and ($\mathbf{W}^K \in \mathbb{R}^{d \times d_k}$), with $d_k=d_q$. Consequently, we derive the keys, values, and queries by projecting $\mathbf{X}$ through the aforementioned matrices, resulting in $\mathbf{Q}=\mathbf{X} \times \mathbf{W}^Q$, $\mathbf{K}=\mathbf{X} \times \mathbf{W}^K$, and $\mathbf{V}=\mathbf{X} \times \mathbf{W}^V$. From a mathematical standpoint, the attention mechanism can be conceptualized as a function mapping input queries $Q$ and key-value pairs $(K, V)$ to a weighted sum of values, facilitating a comprehensive contextual comprehension.
The output $\mathbf{Z}\in\mathbb{R}^{n \times d_v}$ of the self-attention layer is:
\begin{align}\label{eq:Attention}
\mathbf{Z}&= \mathbf{softmax}\left(\frac{\mathbf{Q}\mathbf{K}^T}{\sqrt{d_q}}\right)\mathbf{V}.
\end{align}
\begin{figure}[!ht]
    \centering
    \includegraphics[width=0.7\linewidth]{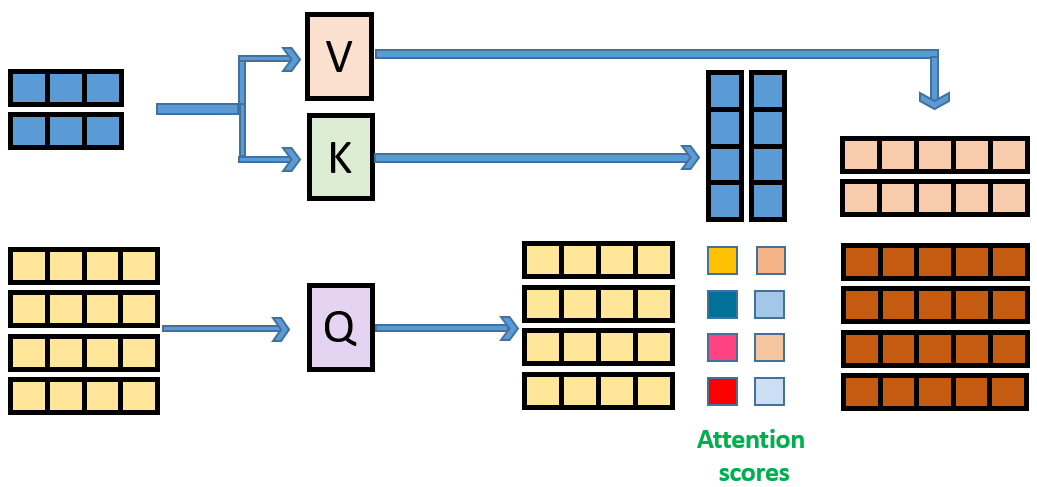}
    \caption{Illustration of the cross-attention operation. 
    }%
    \label{fig:cross_attention}
\end{figure}
The \textbf{Masked attention layer}, a crucial component within Transformer models, introduces a nuanced approach to self-attention. 
Consider the Transformer model described by Vaswani et al.~\cite{vaswani2023attention}, which is often trained to predict the next entity in a sequence. In such cases, it's imperative to prevent the model from "cheating" by peaking into future entities during training. This necessity gave birth to the concept of masking within the self-attention mechanism. This is achieved through a simple yet powerful element-wise multiplication operation with a mask matrix $\mathbf{M} \in \mathbb{R}^{n \times n}$. The mask matrix $\mathbf{M}$ is typically an upper triangular matrix, ensuring that each element in the sequence can only attend to preceding or current elements, but not to future ones.
Mathematically, despite the masking, the attention mechanism retains its core functionality. It still serves as a function that maps input queries $Q$ and key-value pairs $(K, V)$ to a weighted sum of values, allowing the model to develop a holistic contextual understanding of the sequence.
Therefore, the output $\mathbf{Z}\in \mathbb{R}^{n \times d_v}$ of the self-attention is defined by:
\begin{align}\label{eq:Attention1}
\mathbf{Z}= \mathbf{softmax}\left (\frac{\mathbf{Q}\mathbf{K}^T}{\sqrt{d_q}} \circ \mathbf{M}\right )
\end{align}
where $\circ$ denotes the (element-wise) Hadamard product.

\textbf{Multi-Head Attention (MHA)} Multi-Head Attention doesn't introduce a new operation but rather amalgamates multiple Attention Layers. This amalgamation facilitates the integration of diverse relationships among the elements within a sequence. These Multi-Head Attentions can be applied across various operations described earlier. Essentially, MHA involves projecting these elements into $H \in \mathbb{N}$ distinct spaces, with $H$ representing the number of heads, utilizing learned linear projections denoted as $ { \mathbf{W}^{Q_h},\mathbf{W}^{K_h},\mathbf{W}^{V_h} }_{h=0}^H$. This sequence of triplets allows for the projection of each entry $\mathbf{X}$ (or two entries in the case of the Cross attention layer) onto a composite matrix $\mathbf{Z}^{\mbox{total}}=[\mathbf{Z}_0,\mathbf{Z}1,\cdots\mathbf{Z}{h-1}] \in \mathbb{R}^{n\times h \cdot d_v}$, followed by projection onto a weight matrix $\mathbf{W} \in \mathbb{R}^{(h \cdot d_v) \times d}$.

\textbf{Cross attention layer:} While the standard Self-Attention layer efficiently compares all elements within a single sequence $\mathbf{X}$, a different challenge arises when dealing with two distinct sequences, say $\mathbf{X}_1$ and $\mathbf{X}_2$. Capturing the "long-term" dependencies between such sequences becomes essential, whether it's comparing two texts in different languages or analyzing a sequence comprising both text and image data, as observed in the CLIP model~\cite{radford2021learning}.
To address this need, Cross-Attention Layers (depicted in figure \ref{fig:cross_attention}) come into play, enabling the computation of attention between two disparate sequences. In these layers, queries are generated for the first sequence, denoted as $\mathbf{Q}_1 =\mathbf{X}_1\mathbf{W}^Q$, while the key-value pairs for the second sequence are created as $\mathbf{K}_2=\mathbf{X}_2\mathbf{W}^K$ and $\mathbf{V}_2=\mathbf{X}_2\mathbf{W}^V$. Subsequently, attention is calculated between these two sequences.
The output of the cross-attention layer is then determined by:
\begin{align}\label{eq:Attention2}
\mathbf{Z}= \mathbf{softmax}\left(\frac{\mathbf{Q}_1\mathbf{K}_2^T}{\sqrt{d_q}}\right)\mathbf{V}_2.
\end{align}

\subsubsection{Architectures}

Transformer architectures, initially devised for machine translation tasks~\cite{vaswani2023attention}, have undergone a gradual evolution for applications in vision tasks. As discussed in~\cite{Khan_2022}, Transformer architectures are conceptualized as DNNs equipped with Cross-Attention or Self-Attention layers, enabling them to discern relationships among elements within one or more sequences.
According to findings in~\cite{Khan_2022}, Transformer architectures for vision tasks can be categorized into three main types:
Uniform-Scale Vision Transformers, Multi-Scale Vision Transformers, and Hybrid ViTs with Convolutions.

\textbf{Uniform-scale Vision Transformers:} 
These techniques operate on a single sequence with a fixed size that remains constant over time. An example of such a technique is the Vision Transformer (ViT) model~\cite{dosovitskiy2021image}, where the Self-Attention layer is consistently applied to a uniform sequence, typically representing a sequence of image patches. To adapt the Transformer architecture~\cite{vaswani2023attention} to the grid structure of images, ViT first breaks down the input image into a sequence of patches, typically arranged in a $16 \times 16$ grid, as elaborated in~\cite{dosovitskiy2021image}. These patches are flattened and transformed into a vector whose dimensionality varies depending on the employed ViT variant. Subsequently, multi-head self-attention is applied iteratively, followed by the addition of a class token for classification purposes, as depicted in Figure~\ref{fig:vit-figure}.
Another notable Transformer model in the vision domain is the Data-efficient image Transformer (DeiT)~\cite{touvron2021training}, which utilizes distillation techniques~\cite{hinton2015distilling} to mitigate the reliance on vast amounts of training data. DeiT maintains the foundational ViT architecture while incorporating several modifications. Notably, it introduces a distillation token during training with a ResNet teacher model, from which the student Transformer learns. These alterations enable DeiT to achieve performance comparable to ViT while utilizing only a fraction of the pre-training data.

\textbf{Multi-scale Vision Transformers:}
In conventional ViTs, the number of tokens and dimension are fixed across all the network blocks. This static architecture presents limitations, as it fails to capture spatial details at varying scales, a capability inherently present in
CNNs. Additionally, these architectures are often very cumbersome.
Research has shown that adopting a multi-stage hierarchical design can address these limitations effectively. In this approach, the number of tokens is gradually reduced while the dimension of token features is progressively increased, resulting in the generation of efficient features for dense prediction tasks~\cite{wang2021pyramid}. These models typically excel in recognition tasks. The essence of these architectures lies in sparsifying tokens by merging neighboring tokens and projecting them into a higher-dimensional feature space. Notable examples of multi-stage ViTs include Pyramid ViT~\cite{wang2021pyramid}, Swin Transformer~\cite{liu2021swin}, and Segformer~\cite{xie2021segformer}.
Swin, also known as Shifted Windows~\cite{liu2021swin}, addresses the challenge of computational efficiency. To achieve this, it introduces sliding windows into the Transformer architecture, akin to those utilized in CNNs. Initially, small patches are employed in the first Transformer layer, which are then progressively merged into larger patches in deeper layers. The transform blocks locally compute attention within windows of $2 \times 2$ patches, utilizing an alternating staggered window configuration to enable the model global input information gathering. 
This effectively reduces
computational complexity to a linear level while preserving detailed processing capabilities.

\textbf{Hybrid ViTs with Convolutions:}
CNNs excel at capturing fine-grained low-level features in images~\cite{zeiler2014visualizing,lecun2015deep,simonyan2014very,szegedy2015going} and have been integrated into various hybrid designs of Vision Transformers (ViTs), particularly in the initial stages for segmenting and tokenizing input images. Several studies~\cite{wang2021pyramid,Yuan2021TransAnomaly,wang2022anodfdnet,mathian2022haloae} have explored the fusion of Transformer-based 
networks
with CNNs to leverage the strengths of both architectures.

For a more comprehensive overview of Transformers, we encourage readers to consult~\cite{Khan_2022}.

\subsection{Foundation Models}
\label{subsec:foundation}
\subsubsection{Definition}

The term "deep learning" was proposed around 2010~\cite{8187120} and later redefined in 2015~\cite{lecun2015deep}. Deep learning refers to a domain of machine learning where the models are Deep Neural Networks (DNNs), trained on larger datasets to accomplish a specific task. The "deep" (related to the depth of the DNN) aspect stems from the increased number of layers within neural networks compared to earlier versions predating 2010. This depth enables capturing more invariant and robust representations, contributing to DNNs versatility across various fields. Over the years, DNNs have witnessed significant advancements, permeating almost every research domain within computer science and more and enabling them to tackle increasingly diverse applications.

Foundation models, a term originating from~\cite{bommasani2021opportunities}, emerged within the Natural Language Processing (NLP) community. These models, typically based on Transformer architectures, differ from standard DNNs by their ability to \textit{transfer} \textit{knowledge} across \textit{multiple tasks} without requiring task-specific annotations.


This capability
stems from training on vast datasets~\cite{schuhmann2022laion5b}, employing various training techniques, often including self-supervised pre-training~\cite{devlin2019bert, openai2024gpt4}. Such training strategies endow these models with remarkable transferability, enabling them to be effortlessly applied to a wide array of downstream tasks with minimal additional effort. Consequently, they represent a paradigm shift in AI development, offering unprecedented versatility and efficiency in solving diverse problems.

\subsubsection{Architectures}

Creating a foundation model from scratch demands substantial resources~\cite{bigscience_workshop_2022}, encompassing both computational power and access to extensive datasets. Early exemples of such models emerged in the realm of language processing, including BERT~\cite{devlin2019bert} and OpenAI's GPT-n series~\cite{openai2024gpt4,brown2020language,radford2019language}. These foundation models have since undergone specialization for a myriad of applications, spanning from image generation~\cite{Oppenlaender_2022, rombach2022highresolution, ramesh2021zeroshot} to vision-language tasks~\cite{radford2021learning} and even domains like music~\cite{copet2023simple} and robotics~\cite{brohan2023rt2}.


Notable open-source foundation models that have gained significant attention include CLIP (Contrastive Language-Image Pre-training) by OpenAI~\cite{radford2021learning} and, to a lesser extent, the Segment Anything Model (SAM) by Meta AI~\cite{kirillov2023segment}.

Unlike traditional task-specific models, CLIP jointly learns images and text via contrastive pre-training. Using large-scale datasets, it associates images with textual descriptions, allowing broad task generalization without task-specific fine-tuning.
Such as zero-shot image classification~\cite{novack2023chils}, image captioning~\cite{zeng2024meacap}, visual question answering~\cite{song2022clip} and most importantly for this survey, anomaly detection~\cite{jeong2023winclip,deng2023anovl,chen2023aprilgan,Li_2024_WACV}. Showcasing its versatility and robustness in multi-modal understanding.
During unsupervised pre-training, CLIP encodes images and captions into a shared latent space using contrastive loss. This breakthrough bridges the semantic gap between language and vision, paving the way for more sophisticated multimodal setups, enabling efficient solutions to complex problems.


SAM is a zero-shot image segmentation framework that generalizes across diverse tasks without task-specific fine-tuning. Trained on massive datasets, it generates accurate, flexible object masks from prompts like points, boxes, or text. SAM's versatility extends to medical imaging~\cite{zhang2023customized}, autonomous driving~\cite{cai2024crowd}, and anomaly detection~\cite{peng2024sam,li2024sam,li2025clipsam,baugh2023zeroshot,cao20232nd}, demonstrating its adaptability across diverse applications.. With a powerful mask decoder and attention mechanism, SAM adapts to new domains, marking a paradigm shift in computer vision. Unifying segmentation under a generalist approach, much like CLIP unified vision and language. Its robust capabilities enable scalable, efficient anomaly detection across applications.

In subsequent sections, we delve into the utilization of these models for anomaly detection tasks.
For more detailed insights into the diverse foundation models, we refer readers to the relevant surveys~\cite{zhou2023comprehensive,zhao2023survey,zhang2023complete}.

\section{Anomaly, Out-of-Distribution, Open-set and Novelty Detection}

\label{section:Defs}

Various disciplines address the challenge of \emph{detecting unknowns}, each with distinct terminology and problem formulations. Anomaly detection, out-of-distribution detection, open-set recognition, and novelty detection have gained attention for their practical significance. Despite differing names, these fields share common ground, defining a specific \emph{in-distribution} and identifying \emph{out-of-distribution} instances within an open-world framework.\\
However, differences emerge in their interpretation of \emph{unknown} (which is often intertwined with their respective evaluation protocols) and the desired output.
While preparing this article, it became evident that few works comprehensively address all these fields simultaneously. Additionally, different communities tend to emphasize varying aspects of the problem, leading to overlapping definitions and interchangeable usage of terms.

Inspired by these surveys \cite{yang2024generalized,hojjati2024self,chalapathy2019deep}, this section aims to offer succinct and precise definitions for each of these concepts, facilitating a clearer understanding of their distinctions.

\textbf{Open-Set Recognition (OSR)} involves the classification of a specific subset of $k$ known classes out of a total of $n$ classes present in a dataset. In this context, the  $n - k$ classes not included in the dataset form the \emph{open-set distribution}. The primary objective of OSR is to accurately classify the $k$ known classes while also identifying the remaining $n-k$ unknown classes as part of the open-set.

\textbf{Novelty Detection (ND)} represents a specialized instance of OSR where $k = 1$, typically referred to as one-class classification. In ND, all training samples belonging to the known class are considered normal and semantically similar. Conversely, the remaining $n - 1$ classes are designed as novel and must be detected as such. It's noteworthy that although ND is characterized by $k = 1$, it can encompass various classes in a one-class classification setting.

\textbf{Out-Of-Distribution (OOD)}
Out-of-Distribution (OOD) detection is a multi-class problem where multiple categories that constitute the training set are treated as In-Distribution (ID) data, akin to the approach in OSR. OOD data represent a shifted distribution to ID data, differing in label or semantic spaces.
The goal is to identify and flag OOD samples while correctly classifying ID samples into their respective classes.

\textbf{Anomaly Detection (AD)}
assumes a dataset with a well-defined concept of normality, requiring a clear definition of what constitutes a normal example for each class. Anomalies are instances that deviate from this predefined notion.
The definition of "normal" must be precise and context-dependent. For example, in a business selling "non-compliant nut detector", normality would mean undamaged, shelled nuts. Any deviation—such as cracked shells or entirely different objects—is classified as an anomaly.
Unlike earlier approaches, samples resembling the training set are not considered normal. As a result, AD operates typically on a per-class basis.


To sum up, the definitions of AD, ND, OSR, and OOD detection reveal both similarities and differences.
AD and ND act as binary classification algorithms, distinguishing between normal and anomalous instances, while OSR and OOD detection tackle multi-class problems, requiring more supervision.
The primary disparity between AD and ND lies in their evaluation protocols. ND identifies novelties across different label spaces, whereas AD operates within a single label space, aiming to detect anomalies—samples within the same label space that exhibit undesirable properties.
On the other hand, OOD detection extends the concept of unknowns to encompass a broader range of learning tasks compared to OSR. Furthermore, OOD detection and AD implicitly assume the rarity of unknowns in their evaluation protocols, whereas OSR and ND do not rely on such assumptions.

For the remainder of this survey, we will focus solely on the anomaly detection field.





\section{Taxonomy \& Outline}
\label{sec:taxonomy}
Initially dominated by CNNs and generative models, VAD methods have evolved significantly over the past decade. Early works employed autoencoders and GANs to learn compact representations of normal events and detect deviations as anomalies \cite{hasan2016learning, gong2019memorizing, ravanbakhsh2017abnormal}. More recent CNN-based frameworks have introduced causality-inspired representations \cite{liu2021learning}, hybrid reconstruction-prediction approaches using diffusion models \cite{wu2023ddhybrid}, local-global modeling techniques \cite{zhao2025rethinking}, and spatial-temporal modeling techniques like CRCL \cite{yu2022crcl} and STNMamba \cite{yu2024stnmamba}. In parallel, efforts have also been made to optimize inference efficiency without compromising accuracy, such as EfficientAD \cite{batzner2024efficientad}. These methods have proven effective in specific domains, yet their reliance on local receptive fields and limited semantic modeling has constrained their adaptability to complex and diverse anomaly scenarios.

In contrast, Transformer-based architectures enable global attention and richer context modeling, which has led to a surge of interest in their application to VAD. Foundation models such as CLIP \cite{radford2021learning} and SAM \cite{kirillov2023segment} further extend this capacity by enabling zero-shot and multimodal understanding. In this survey, we focus on unsupervised approaches leveraging these architectures, which we argue represent a paradigm shift in VAD research. While we center on these newer directions, we also briefly acknowledge the classical models to contextualize this progression.

VAD primarily relies on unsupervised or self-supervised learning (SSL), using only normal samples for training. These paradigms focus on identifying irregularities by capturing and modeling the intrinsic characteristics of normal data. In some cases, synthetic anomalies are generated as a form of supervisory signal, exemplified by techniques such as CutPaste~\cite{li2021cutpaste}.

\sloppy
Unsupervised VAD methods can be broadly categorized into three main families: reconstruction-based methods, feature-based methods, and zero-/few-shot anomaly detection (AD). This division reflects the different ways anomalies are detected, either through data fidelity (reconstruction) or semantic representations (features). Transformers have emerged as a transformative tool across these approaches, offering significant advantages in performance and adaptability. Their strong pre-training capabilities make them exceptionally well-suited for unsupervised and SSL, while their sequential structure and compatibility with prompting provide flexibility for diverse data pipelines and tasks.



\begin{figure}
    \centering
    \includegraphics[width=0.99\linewidth]{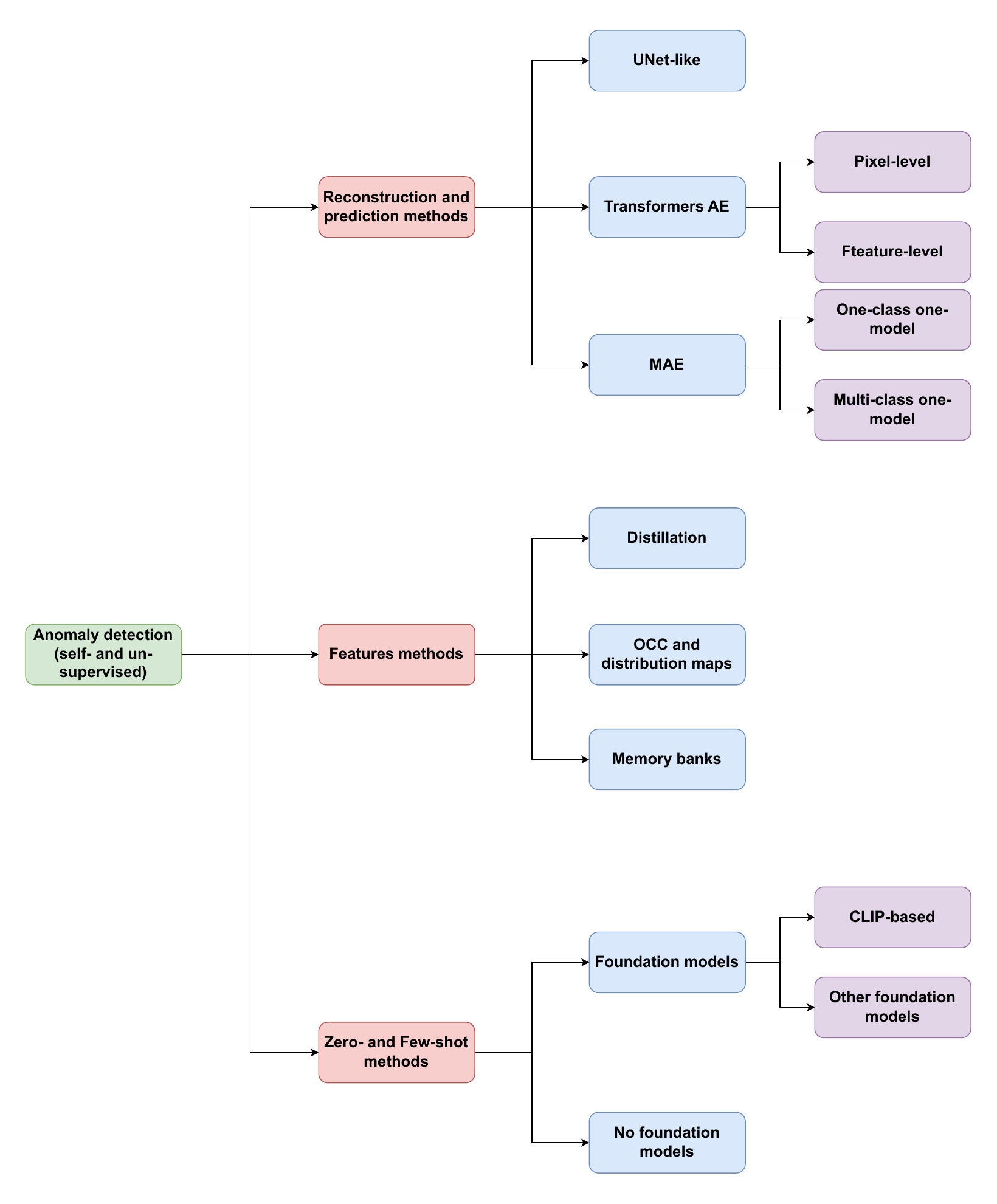}
    \caption{Taxonomy of anomaly detection methods.}
    \label{fig:taxonomy}
\end{figure}

This survey explores the integration of Transformers into VAD, examining their contributions and challenges across various methods. 
 We analyze reconstruction-based approaches by their architectures and feature-based methods by their anomaly detection strategies.
Figure \ref{fig:taxonomy} shows the defined taxonomy for this survey.
Additionally, we discuss zero- and few-shot detection, emphasizing foundation models impact on their rise. Our goal is to provide a comprehensive view of Transformers influence on VAD and their potential for advancing anomaly detection. 

While many efficient variants of transformers have emerged \citep{chen2021mobilevit,mehta2022mobilevitv2,li2023mobileformer}, specifically tailored for edge devices. Their use in unsupervised VAD is still scarce and emerging, future work will likely expand on this trend by optimizing pre-trained Transformers for deployment under compute and memory constraints.

\setlength\tabcolsep{0pt}
\begin{table}[h!]
    \small
    \renewcommand{\arraystretch}{1.6} 
    \renewcommand{\cellalign}{vh}
\renewcommand{\theadalign}{vh}
    \setlength{\tabcolsep}{10pt} 
    \resizebox{1\textwidth}{!}{
    \begin{tabular}{|m{2.7cm}|m{3.1cm}|m{1.8cm}|m{10cm}|} 
        \hline
        
      \rowcolor{lightgray} \textbf{Paradigm} &\multicolumn{2}{|l|}{\textbf{Approach}} & \textbf{Methods} \\
        \hline
        \multirow{5}{*}{\vspace{-0.9in}\textbf{Reconstruction}}  
            &  \multicolumn{2}{|l|}{UNet-like} & \cite{Yuan2021TransAnomaly,feng2021convolutional,yang2024attention,wang2022anodfdnet,luo2023normal,zhang2023video,he2024video,taghinezhad2023new,aslam2024demaae,aslam2023vae,aslam2022a3n,le2023attention,UZEN2022118269,chen2022utrad,ZHOU2024114216,xiao2024anomaly,zhao2023omnial,liu2021sagan,yang2022transformer,yang2023video}   
            \\ \cline{2-4} 
  
            & \multirow{2}{=}{\vspace{-0.2in}Transformer AEs} & Pixel-level & \cite{Mishra_2021,article,shang2023defect,lee2022anovit,venkataramanan2020attention,liu2020visuallyexplainingvariationalautoencoders,pinaya2022unsupervised,ZHANG202357,aslam2024transganomaly,10466765}\\   \cline{3-4}
            &  & Feature-level & \cite{mathian2022haloae,yao2022generalizable,CAI2023106677,zhang2023exploring,you2022adtr,yao2023visual,yang2024context,wang2024transformer,LUO2024107810,yao2024priornormalityprompttransformer,lu2023hierarchical,huang2022unsupervised,liang2023omni} \\   \cline{2-4}
            & \multirow{2}{*}{\vspace{-0.2in}MAE} & One-class One-model& \cite{pirnay2021inpainting,De_Nardin_2022,rashmi2024ano,yao2022siamese,tian2023unsupervised,fu2024spatiotemporal} \\  \cline{3-4}
            &  &  Multi-class & \cite{you2022unified,KANG2024111186,Yao_Zhang_Li_Sun_Liu_2023,DBLP:journals/access/ChenPHZJ23,lee2022multicontextual,ristea2023selfdistilled,ristea2022selfsupervised,Madan_2024} \\  
        \hline
        \multirow{3}{=}{\vspace{-0.2in} \centering\textbf{Features}}
        &  \multicolumn{2}{|l|}{Distillation}    & \cite{cohen2022transformaly,baradaran2023multi,yao2023learning,Bozorgtabar_Mahapatra_2023,zhang2024feature,li2024sam,zhang2023faster,zhang2023destseg,xiang2023squid,gong2019memorizing}\\   \cline{2-4} 
        &  \multicolumn{2}{|l|}{OCC and Distribution Maps}   & \cite{cmc.2023.035246,visapp23,barbalau2023ssmtl++,chang2022video,yan2022cainnflow,silva2023attention,PangFusion,tailanian2024u,yao2024local,chiu2023self,yao2023focus,Bozorgtabar_2022_BMVC,wu2022self,li2024cross,liu2025crcl,liu2024ampnet}\\   \cline{2-4} 
        &  \multicolumn{2}{|l|}{Memory Banks}    & \cite{10007829,lee2023selformaly,gong2019memorizing,liu2023component,wang2023multimodal,pang2022masked}\\  
        \hline
         \multirow{5}{=}{ \textbf{Zero- \&  Few- Shot }} 
            & \multirow{3}{*}{ \vspace{-0.3in} Foundation  Models} & CLIP \& SAM  & \cite{jeong2023winclip,baugh2023zeroshot,cao20232nd,chen2024clip,zhu2024llms,tamura2023random,chen2023aprilgan,deng2023anovl,zhou2023anomalyclip,Li_2024_WACV,cao2025adaclip,gu2024filo,li2025clipsam,kim2023unsupervised,zhu2024toward,li2024promptad,peng2024sam} \\ \cline{3-4}
            &  & Other Foundation Models & \cite{gu2023anomalygpt,li2023myriad,zanella2024harnessing,yang2025follow,chen2023tevad,zhang2023exploringVQA,cao2023towards} \\ \cline{2-4}
            &  \multicolumn{2}{|l|}{Few-shot without Foundation Models} &\cite{Takimoto2022,9711445,huang2022registration,fakhry2024enhancing,li2024musc,schwartz2022maeday} \\ \cline{2-4}        
        \hline
    \end{tabular}}
    \caption{Representative algorithms for each sub-category in this survey.
    }
    \label{tab:anomaly-detection}
\end{table}

\section{Reconstruction and Prediction Based}
\label{sec:reconstuction_based}

Reconstruction-based approaches predominantly use the auto-encoder (AE) framework, detecting anomalies by comparing input data with its reconstructed output.
The core principle behind this approach is the generalization gap assumption: models trained on normal samples excel at reconstructing normal regions but struggle with anomalous regions due to the distributional shift.

A key challenge is the identity mapping trap: instead of learning meaningful variations within the data, the model defaults to a trivial identity function, reconstructing both normal and anomalous regions with high fidelity. 
As noted by \citet{gong2019memorizing}, this behavior often results from overfitting, worsened by the limited size and diversity of anomaly detection datasets. Overcoming this issue is crucial for the effectiveness of these methods.

A key differentiator among these methods is how they mitigate the identity mapping trap, primarily through reconstruction network design and reconstruction target selection.
Recent advances integrate attention mechanisms and Transformers, improving the capture of complex patterns among other potential benefits.
This section explores these developments, categorizing methods by architectural design, including UNet-like structures, standard auto-encoders, and masked auto-encoders.


Prediction-based methods forecast future states, while reconstruction-based methods recreate inputs. Despite this distinction, both share similar anomaly scoring strategies and architectural foundations, our key criterion for categorizing reconstruction methods.
Their overlapping designs and reliance on learned normal data representations justify grouping them together to highlight shared principles and implementation strategies.


\subsection{UNet-Like Architectures}\label{subsec:UNet_AE}


Early efforts to integrate Transformers into reconstruction-based anomaly detection relied on CNNs, particularly the UNet architecture~\cite{ronneberger2015unet}, a symmetric encoder-decoder with skip connections for multi-scale feature fusion. Over time, attention mechanisms were gradually incorporated to address its AD-specific limitations, eventually leading to fully Transformer-based designs.

\begin{figure}[!ht]
    \centering
    \includegraphics[width=0.99\linewidth]{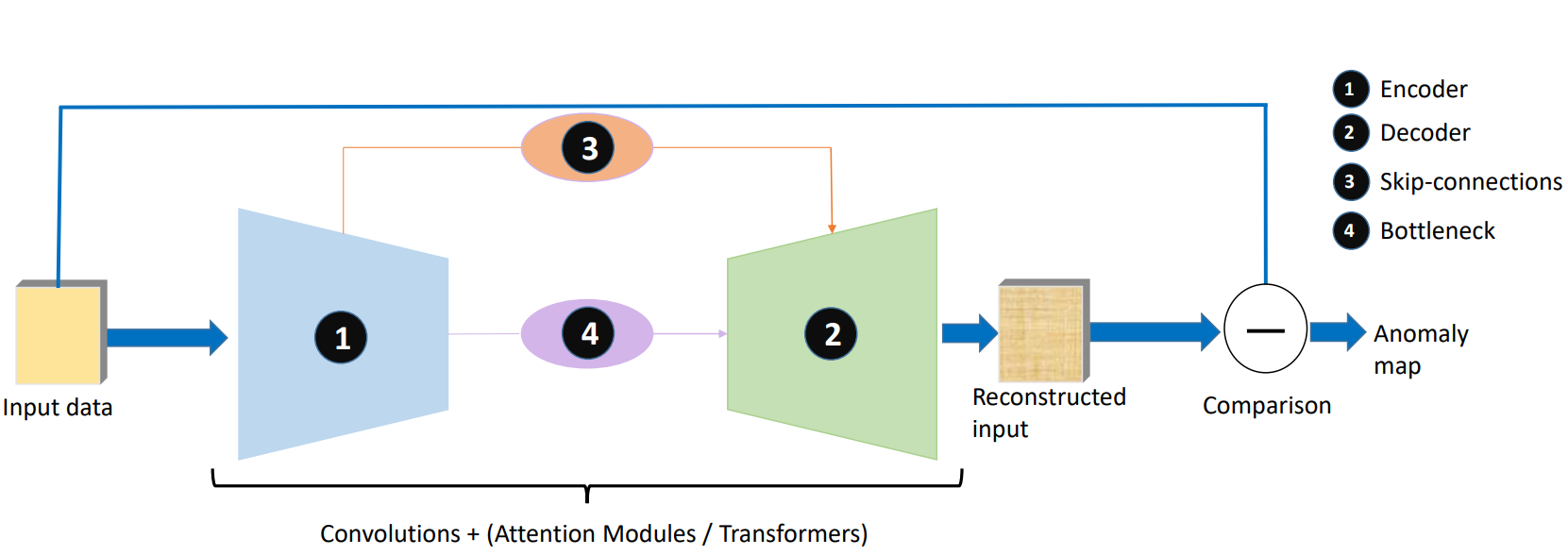}
    \caption{Overview of UNet-like reconstruction- and prediction-based AD methods. Transformers/Attention can be integrated into four main components: Encoder, Decoder, Skip Connections and Bottleneck. Anomalies are detected by comparing the reconstructed output with the input to generate an anomaly map.}

    \label{fig:UNet_fig}
\end{figure}

This subsection traces this evolution, from hybrid models integrating attention into CNNs to fully Transformer-based UNet-like architectures. The discussion follows their architectural progression, starting from encoder-decoder bottlenecks and expanding to other key algorithmic details. Figure \ref{fig:UNet_fig} visually summarizes these approaches.

The bottleneck serves as a filter enforcing structured data representation to highlight informative patterns. Many studies have enhanced this component by integrating attention mechanisms and Transformers.
Several approaches focusing on this aspect have been proposed for video, \citet{Yuan2021TransAnomaly} incorporates a ViViT~\cite{arnab2021vivit} at the bottleneck, leveraging its spatiotemporal feature extraction capabilities. \citet{feng2021convolutional} integrates attention at the bottleneck by extending MHSA to the temporal dynamics. Additionally, a dual discriminator GAN is used to enforce local frame-level consistency and global temporal coherence.
Expanding on these ideas, \citet{yang2024attention} uses a pseudo-anomaly generation module for training, coupled with channel and spatial attention \cite{hu2019squeezeandexcitation} to generate attention maps highlighting frame foreground. These are added to the encoder's features to reduce background interference during reconstruction.


Subsequently, several approaches have included a template/prototype component, i.e. utilizing a reference of normality from the training data and contrast it with the input. This aids to filter-out anomalous signals at the bottleneck before reconstruction, thereby mitigating identity mapping. 
\citet{wang2022anodfdnet} employs a reference normal sample as a template, contrasting image features against this reference within a ViT-based bottleneck.
Similarly, \citet{luo2023normal} includes attention-based bottleneck with a two-stage training process. The first stage focuses on feature learning, after which the encoder is frozen. In the second stage, the bottleneck is fine-tuned to suppress anomalous feature signals using synthetic anomalies and a reference template. 

For video anomaly detection, prototype-based approaches usually operate at the feature level. For instance, \citet{zhang2023video} uses a dynamic prototype unit~\cite{lv2021learning} to learn and aggregate prototypes in real-time with encoder features. \citet{he2024video} refines feature representations using the Convolutional Block Attention Model (CBAM)\cite{woo2018cbam} before storing them in a memory bank (see Section~\ref{subsec:memory-bank}) for inference-stage comparisons. Meanwhile, \citet{taghinezhad2023new} directly queries a multi-scale memory bank via an attention module.

Integrating Transformers and attention mechanisms at the encoder-decoder level enhances representational and reconstruction capabilities. Additionally, attention-augmented skip connections regulate the flow of multi-scale information between these two components.

Several video AD approaches leveraged these benefits, by including attention-based mechanisms at the decoder level.
DeMAAE \cite{aslam2024demaae} incorporates a global attention module within the CNN and ConvLSTM layers~\cite{medel2016anomaly}, utilizing hidden states to compute attention maps that refine reconstruction.
A-VAE~\cite{aslam2023vae} employs an attention-enhanced VAE, combining CNNs, Bi-LSTMs, and multiplicative attention mechanisms in the decoding phase to emphasize discriminative spatiotemporal features.
Meanwhile, dual stream approaches are utilized in~\cite{aslam2022a3n,le2023attention}.
\citet{le2023attention} uses a lightweight dual-stream encoder focusing for spatial and temporal feature extraction. The attention-based decoder integrates a cascade of channel attention layers before applying residual connections. 
\citet{aslam2022a3n} combines attention with an adversarial AE with a symmetric two-stream decoder for frame reconstruction and future frame prediction. Both streams use global attention mechanisms, with an adversarial training strategy in the predictive stream.

For image-level AD, \citet{UZEN2022118269} utilizes a pre-trained inception-v3~\cite{szegedy2015rethinking} encoder with a Swin decoder. The multi-scale decoder features are fused using a squeeze-excitation~\cite{hu2019squeezeandexcitation} based attention module.

 Other image-level methods explored the integration of attention mechanisms in both the encoder and decoder.
 \citet{ZHOU2024114216} replaces UNet's convolutional blocks with convolutional Transformers. It employs depth-wise separable convolutions to compute MHSA while fusing multi-scale features at each skip connection.
 UTRAD~\cite{chen2022utrad} substitutes CNN blocks with ViTs for feature reconstruction. Skip connections transmit both class tokens and query embeddings, with an integrated bottleneck to regulate information flow and reduce identity mapping risks. \citet{xiao2024anomaly} proposes an in-painting approach (see Section~\ref{subsec:MAE}) using a ConvNeXt-based UNet~\cite{liu2022convnet}. The model incorporates squeeze-and-excitation channel attention at each UNet block, while a memory bank filters out potential anomalies. 
 OmniAL~\cite{zhao2023omnial} tackles the complex task of multi-class anomaly localization (see section~\ref{Multi-class}) with a framework comprising three key components: (1) a panel-guided anomaly synthesis network, inspired by \citet{wu2013just} generates synthetic anomalies for training; (2) a reconstruction sub-network equipped with dilated channel and spatial attention (DCSA) at every block is 
 learned
 to map synthesized anomalies back to normality; (3) A UNet-like localization sub-network, integrating the DSCA blocks, refines anomaly localization.

Similar to reconstruction approaches, generation methods have significantly benefited from the integration of attention mechanisms, particularly at skip connections.
SAGAN~\cite{liu2021sagan} extends Skip-GANomaly~\cite{akçay2019skipganomalyskipconnectedadversarially} by incorporating a Convolutional Block Attention Model within the GAN's generator skip connections. This enhances the feature maps global semantics, improving anomaly detection capabilities.
AnoTrans~\cite{yang2022transformer} refines SAGAN by introducing self-attention at skip connections, capturing localized features more effectively.
Moreover, AnoTrans adopts a fully Transformer-based architecture, replacing traditional CNN blocks with Swin blocks integrating patch merging and expansion layers to mimic UNet's feature maps scale variations.

For video anomaly detection, \citet{yang2023video} proposes a similar U-shaped Swin architecture with dual skip connections for key-frame-based restoration. The skip connections include a cross-attention path and a temporal upsampling path, allowing the model to reconstruct intermediate frames from key frames. Without explicit motion cues, the model is optimized for intrinsic temporal continuity, enhancing its high-level semantics while reducing the risk of identity mapping.

\subsection{Transformers Auto-Encoder}\label{subsec:Transformers_AE}

Unlike UNet, the auto-encoder (AE) architecture omits skip connections, relying on an encoder, decoder, and a feature fusion network. Typically trained in an unsupervised or self-supervised manner, AEs learn compact representations of normal data, ensuring high reconstruction fidelity for normal regions while revealing anomalies through reconstruction or prediction errors.

\begin{figure}[!ht]
    \centering
    \includegraphics[width=0.99\linewidth]{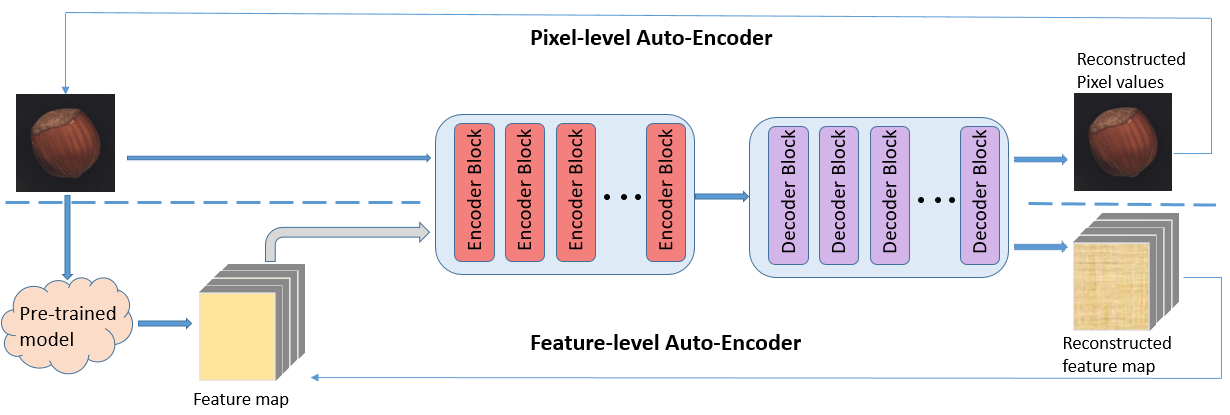}
    \caption{The basic flow of auto-encoder reconstruction and prediction based AD methods.
The methods differ primarily in their reconstruction targets: pixel-level methods (top) reconstruct the full image, while feature/token-level methods (bottom) use pre-trained features/tokens as the reconstruction target.}%

    \label{fig:AE_Reconstruction}
\end{figure}

Transformers, are well-suited to these training paradigms~\cite{caron2021emerging,wang2022self,he2021masked}, present a promising avenue for enhancing AE-based anomaly detection with strong representation learning and global receptive fields. AE methods primarily differ in their reconstruction objectives (Figure \ref{fig:AE_Reconstruction}): pixel-wise approaches reconstruct the full input, while feature-level methods use tokenized pre-trained features as reconstruction targets, 
thus leveraging the semantic and contextual information of pre-trained models while introducing an additional robustness to noise.

\subsubsection{Pixel-Level AE}

 A common way to leverage Transformers strong representation capabilities and global receptive field is to use them as encoders. This has been leveraged in VT-ADL~\cite{Mishra_2021}, ViV-Ano~\cite{article}, DAT-Net~\cite{shang2023defect} and AnoViT~\cite{lee2022anovit}, all using a ViT encoder with a CNN decoder, while adding feature modeling components to enhance performance.
AnoViT rearranges the encoder's patch embeddings for reconstruction, while VT-ADL trains an additional GMM on the learned embeddings for localization. 
\citet{shang2023defect} enhances the ViT encoder’s spatial awareness by adding two modules: (1) A local dependencies modules using K-nearest neighbor and MLP layers, enriching patch embeddings with local information; (2) A dynamically defined auxiliary graph positional encoding, facilitating the integration of structured relationships among image patches.
ViV-Ano~\cite{article} leverages ViTs representations to train a VAE, mitigating the data requirements of Transformers while leveraging their robust representations. 
Beyond ViTs, \citet{ZHANG202357} uses a Swin encoder with a mutual attention-based feature fusion module, enhancing local patch information integration within the encoder's representation.

Other studies have explored the integration of VAEs with attention mechanisms. \citet{venkataramanan2020attention} integrates attention at the loss function to refine latent VAE normalcy representations. \citet{liu2020visuallyexplainingvariationalautoencoders} employs a gradient-based attention mechanism in the latent space to enhance localization accuracy. \citet{pinaya2022unsupervised} combines 
VQ-VAE with Transformers, using an ensemble of auto-regressive Transformers to model the compact representations learned by VQ-VAE, filtering out low-likelihood anomalies before reconstruction to prevent identity mapping.

In video anomaly detection, TransGANomaly~\cite{aslam2024transganomaly} merges GANs adversarial training with Transformers. A ViViT-based predictive generator is used for next-frame prediction, reinforced by overlapping 3D frame tokens for temporal continuity, alongside a CNN-based discriminator. Similarly, \citet{10466765} uses future frame prediction, with a ViT-based spatio-temporal encoder for aerial drone footage, utilizing temporal cross-attention between class tokens to aggregate temporal features effectively.



\subsubsection{Feature-Level AE}
Previous studies have shown that normal and abnormal samples are effectively separable in the feature space of pre-trained models~\cite{Bergmann_2020}.
Building on this, numerous AE solutions utilizing feature maps as reconstruction targets were introduced.

Early approaches retained some convolutional backbones, while incorporating Transformers to improve representational power, leveraging their advantages for reconstruction.
HaloAE~\cite{mathian2022haloae} employs HaloNet~\cite{vaswani2021scaling}, a hybrid model combining local self-attention with convolutional components.
Similarly, \citet{yao2022generalizable} propose the use of a ViT-based inductive Transformer encoder, learnable auxiliary induction tokens aggregate normalcy semantics. These tokens, appended to the encoder’s input, replace the original features in the reconstruction process, mitigating identity mapping. Additionally, convolution-based learning of $(q,k,v)$ attention triplets improves data efficiency. \citet{CAI2023106677} also use an inductive Transformer encoder with a deconvolution-based decoder. An additional MLP layer estimates feature variance, amplifying anomalous regions reconstruction error.

Expanding on hybrid approaches, subsequent works have focused on fully Transformers-based architectures, primarily using ViTs.
\citet{zhang2023exploring} demonstrate that a plain ViT-based AE can achieve state-of-the-art performance through carefully designed macro and micro architectural adjustments. Major adjustments include employing a single-layer bottleneck, utilizing shallow ViT blocks for segmentation, and self-supervised pre-training of the encoder. Additional refinements, such as removing layer normalization in the final encoder layer and using patch tokens with positional embeddings instead of class token for reconstruction, further optimize performance.
On a similar note, ADTR \cite{you2022adtr} highlights the effectiveness of query embeddings to address identity mapping. To harness this property, its fully ViT-based AE architecture incorporates learnable query tokens representing normalcy, which are appended to the decoder. \citet{yao2023visual} augment the ViT with a memorial attention mechanism, creating a dual-stream attention-driven reconstruction. Instead of relying on direct reconstruction discrepancies, the method employs a Normalizing Flow (NF)~\cite{dinh2015nice} to model the joint distribution of the two streams.
Addressing multi-class AD using a ViT encoder, RASFormer \cite{yang2024context} introduces a sequential reconstruction process at each decoder block. Thus enhancing contextual and spatial awareness through an adaptive gating mechanism while using an SSL feature-denoising learning objective.

In the video domain, \citet{wang2024transformer} integrate object detection, optical flow analysis, and feature fusion with a static filtering strategy. Object-level features are emphasized to reduce background interference, while static filtering removes fixed objects. Additionally, temporal-order embeddings are added to ViT’s positional embeddings to enhance temporal feature learning.

A key research direction involves incorporating templates or prototypes to induce information loss in anomalous samples, increasing reconstruction errors.
TFA-Ne~\cite{LUO2024107810} and PNPT~\cite{yao2024priornormalityprompttransformer} both leverage normality templates. TFA-Net concatenates reference normal image features to the ViT-based AE input, employing self-attention to suppress anomalous signals, reconstructing the template embeddings afterwards to mitigate identity mapping.
In contrast, PNPT targets multi-class tasks by constructing a category-specific prior-normality prompt pool using pre-trained features. 
These prompts are aligned with the encoder's features via cross-attention to filter anomalies before reconstruction.
With a similar objective, HVQ-Trans~\cite{lu2023hierarchical} adopts a quantized approach for prototype modeling, using a VQ-based codebook~\cite{ramesh2021zeroshot} to store per-category normality prototypes. During reconstruction, encoder embeddings are replaced with most similar prototypes, and a VQ-based Transformer decoder applies cross-attention for refined reconstruction.


Generation-based AE methods derived significant benefits from the integration of attention and Transformers.
\citet{huang2022unsupervised} propose a two-stage patch-generation strategy using a Swin-based AE, where the model generates each patch conditioned on 
its left- and top- predecessors
while processing the image bi-directionally to minimize bias.
Similarly, OCR-GAN \cite{liang2023omni} leverages the distinct frequency distributions of normal and abnormal images by incorporating two specialized components into the GAN architecture. 
First, a frequency decoupling module splits the image into three subsets, each reconstructed independently through parallel branch networks. Second, an attention-based channel selection module enables adaptive feature interaction across branches, improving frequency-specific feature learning and integration.

\subsection{MAE-Like Architectures}\label{subsec:MAE}

Masked Autoencoders (MAEs) \cite{he2021masked}
have showcased Transformers ability to learn highly expressive representations through self-supervision. By masking portions of the input, MAEs force the model to understand global structures from limited information, which mitigates identity mapping.
However, the data scarcity in anomaly detection (AD) poses challenges in directly applying MAEs. Recognizing their potential, researchers have adapted MAEs by modifying their architectures and training strategies.


A notable adaptation is replacing the lightweight decoder used in standard MAEs with a deeper decoder, a critical requirement for precise localization.
Being strong representation learners, other key adaptations aim to constrain the information the MAE learns to avoid capturing anomalous signals, which could lead to identity mapping. 

\begin{figure}[!ht]
    \centering
    \includegraphics[width=0.99\linewidth]{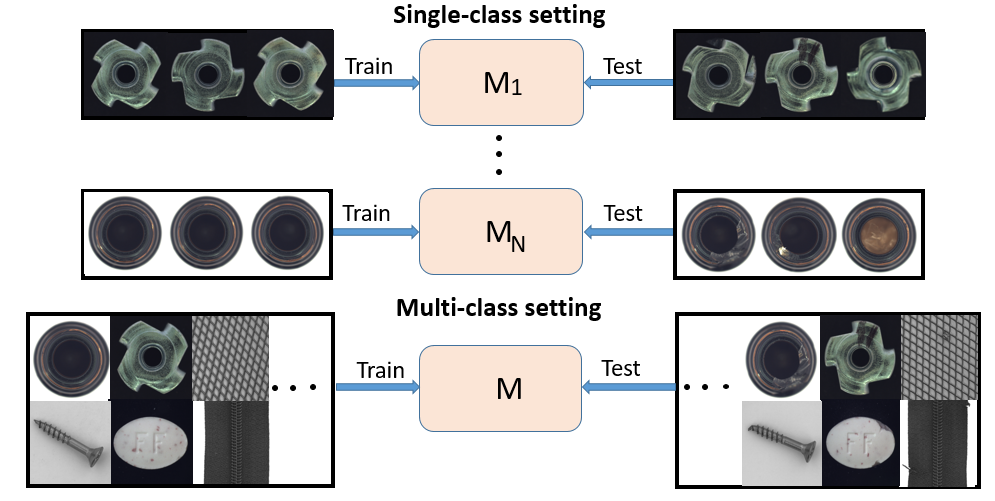}
    \caption{The basic setups for multi-class and single-class anomaly detection. Single-class setup (top) uses a new model for each class in the dataset. The multi-class setting (bottom) uses a single model to perform AD on all classes.}%
    \label{fig:MAE_setup}
\end{figure}

MAEs strong visual representation capabilities make them suitable for multi-class AD, as they can aggregate normality across multiple classes without confusion. Hence, we categorize these methods into one-class and multi-class setups, as illustrated in Figure~\ref{fig:MAE_setup}.


\subsubsection{One-Class One-Model}

Before the emergence of MAE models, \citet{pirnay2021inpainting} explored a similar in-painting strategy for AD, employing a U-Net-like Transformer architecture to reconstruct missing image patches. To mitigate identity mapping, the MHSA operation was modified by projecting the query and key vectors, with reconstruction performed patch by patch.
 
Building on the MAE framework, \citet{De_Nardin_2022} tackles the issue of anomalous signals leaking into unmasked regions embeddings, by reconstructing each patch using only its neighbors. Besides the patch reconstruction, vertical and horizontal image stripes reconstruction is added to increase robustness.
Conversely, \citet{yao2022siamese} introduce a feature decoupling strategy, where pre-trained features are split into two disjoint subsets to serve as the MAE input. A Transformer decoder reconstructs each subset using the complementary set, effectively functioning as a distillation target (see section~\ref{subsec:distillation}).
In medical imaging, Ano-swinMAE~\cite{rashmi2024ano} adapts the MAE framework to data-scarce environments by using a Swin-based MAE with a sliding window masking strategy. A heavy post-processing component refines anomaly maps, significantly improving localization accuracy.

Some methods integrate a memory component to enhance normalcy representations.
MemMC-MAE~\cite{tian2023unsupervised} integrates a memory module in the encoder to filter out anomalous signals before reconstruction. Leveraging this memory, the decoder maximizes similarity with the encoder’s representations via multi-scale cross-attention, ensuring a more reliable reconstruction.
SMAMS~\cite{fu2024spatiotemporal} extends this concept to video data 
using spatiotemporal MAEs and memory modules at the bottleneck to filter out abnormal features.
Video frames are embedded as spatiotemporal cubes while using skip connections for improved reconstruction.

\subsubsection{Multi-Class}
\label{Multi-class}
\citet{you2022unified} highlights the potential of Transformer models in reconstruction-based AD. This study pinpoints a notable Transformers advantage, its query embeddings being less prone to identity mapping than convolution operations.
Building on this, the authors propose a layer-wise query decoder that inserts encoder query embeddings into each decoder layer via cross-attention, incorporating neighborhood masked attention and feature jittering to further reduce identity mapping.
Similarly, MSTAD \cite{KANG2024111186} emphasizes query embeddings role, applying independent masking on the feature map and channel dimensions and using a per-category learnable query embedding as decoder inputs to filter out anomalous embeddings via cross-attention.

Multiple two-stage approaches have been proposed. For instance,
MALA~\cite{DBLP:journals/access/ChenPHZJ23} is a ViT-based two-stage method, which 
first
augments MAE training with a pseudo-label (ground-truth or reconstructed) prediction component in a  generative-adversarial way, which is then discarded. In the second stage, the encoder is frozen to fine-tune the decoder using synthetic anomalies, following the strategy employed in DRAEM~\cite{zavrtanik2021draem}.
One for All~\cite{Yao_Zhang_Li_Sun_Liu_2023} employs a symmetric ViT AE, adopting visual tokens as in DALL-E~\cite{ramesh2021zeroshot} as reconstruction targets. At inference, prototype-guided proposal masking removes potential anomalies before reconstruction. 
Addressing video AD computational challenges, \citet{ristea2023selfdistilled} incorporate convolutional Transformer blocks and motion-gradient-based token weighting. This method optimizes the training pipeline by introducing a second-phase training for a self-distilled decoder, refined using synthetic anomalies for better localization.

Processing video frames as tokens has allowed the extension of MAEs to the video domain with novel approaches. \citet{lee2022multicontextual} propose a dual-decoder MAE, performing frame-masking and optical flow estimation respectively, balancing spatial and motion information reconstruction.
In contrast, \citet{ristea2022selfsupervised} introduce a convolutional MAE block that integrates reconstruction into its architecture. This block combines dilated, centrally masked convolution with a channel-wise Transformer modules, multiplying the activation maps by the resulting attention tokens for reconstruction.
\citet{Madan_2024} refine this approach by replacing standard attention with channel-wise attention and adapting the loss function for better performance.

\subsection{Discussion}
\label{discussion_reconstruction}

Leveraging the generalization gap assumption, reconstruction and prediction-based methods employ a variety of auto-encoders (AEs) architectures and aim to improve their discriminatory abilities for anomaly detection (AD). The widespread adoption of Transformers and attention mechanisms has significantly enhanced the capacity of these methods to model complex patterns, leverage large-scale pre-training and reduce the risk of identity mapping.
Key innovations attributed to Transformers include leveraging query embeddings for AD \cite{you2022unified,KANG2024111186,you2022adtr}, providing higher robustness and ease of adaptation to the multi-class settings through conditional information encoding.
Self-supervised methods, particularly MAEs, allow models to develop a deeper contextual understanding, enhancing robustness against identity mapping traps and facilitating multi-class settings. Additionally, shifting toward reconstructing pre-trained features derived from large-scale pre-training as targets, boosts the algorithms discriminative power \cite{Bergmann_2020,salehi2020multiresolution}.

Despite these advancements, reconstruction-based methods tend to struggle in 
scarce
data regimes due to 
their
extensive training requirements, in addition to their 
limited
scalability for resource-constrained applications. Identity mapping remains a persistent challenge, especially when using pixel-level reconstruction objectives.
Moreover, these methods often struggle to balance global context modeling with fine-grained, low-level feature precision. 
Several open problems remain, including the limited interpretability of these methods, making it difficult to understand how models distinguish between normal and anomalous patterns.
Addressing these challenges is crucial 
to make AD methods really useful and generalizable.

The reconstruction and prediction-based paradigm is likely to continue evolving, driven by advancements in self-supervised learning and hybrid architectures.
The development of hybrid CNN-Transformer architectures and local-aware attention modules holds promise for balancing fine-grained local information with global context modeling.
Improved uncertainty estimation methods could enhance interpretability and decision confidence, while few-shot learning strategies and synthetic anomaly generation may address data scarcity challenges. 
Additionally, solutions incorporating feature-based modeling components (see Section~\ref{sec:features}) offer a promising direction to alleviate reliance on AE architectures.
While the field continues to innovate, addressing these existing challenges will be pivotal to unlocking the full potential of reconstruction and prediction-based methods in AD.

In summary, while Transformers have significantly advanced reconstruction-based VAD, 
addressing
their limitations remains an open challenge, driving further innovation in the field.

\section{Feature Based}
\label{sec:features}
Feature-based methods using CNNs, such as PatchCore~\cite{Roth_2022_CVPR}, SPADE~\cite{park2019semantic} and PADiM~\cite{defard2021padim}, have demonstrated a strong performance while maintaining simplicity. 
Their effectiveness stems from leveraging robust vision models capable of extracting discriminative features that separate normal from anomalous patterns. 
These models act as feature extractors, with methods differing mainly in how they process and partition features into normal and anomalous subsets.
Effective anomaly detection requires features that encode both global and local information while preserving spatial awareness. Hence, with their global receptive field and spatial modeling capabilities, Transformers offer a promising enhancement to feature-based methods.

These methods can be broadly grouped into three main categories: Distillation, distribution map or one-class classification, and memory banks or feature matching methods.


\subsection{Distillation}\label{subsec:distillation}
 Introduced by \citet{hinton2015distilling}, distillation is a knowledge transfer approach where a teacher model, typically pre-trained on a large-scale dataset, guides the training of a smaller, more resource efficient student model.

\begin{figure}[!ht]
    \centering
    \includegraphics[width=0.99\linewidth]{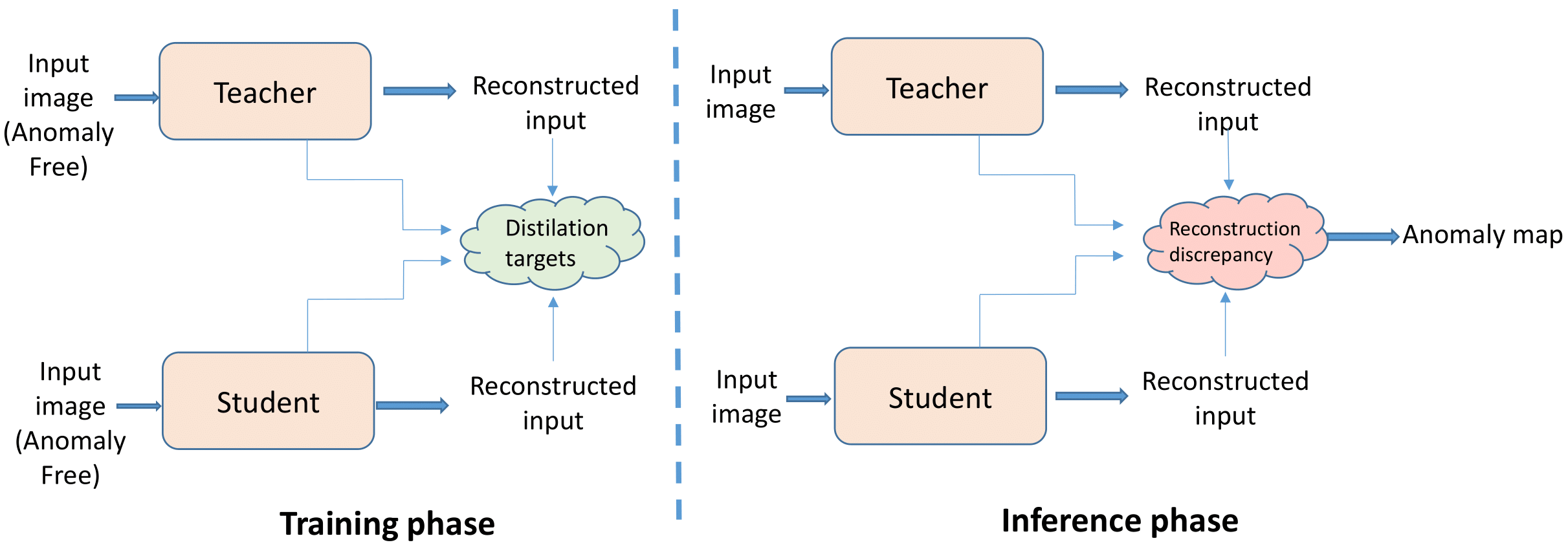}
    \caption{The basic flow of distillation-based methods. For training, the student is guided by the teacher to learn normal representations. At inference, the teacher-student discrepancy is indicative of anomalies, as their representation similarity is constrained to normalcy due to their training.}  
    \label{fig:distillation_setup}
\end{figure}

In the context of AD, the teacher model excels at extracting distinguishable features for normal and anomalous data~\cite{Bergmann_2020,salehi2020multiresolution}. The distillation process, conducted solely on normal samples, is based on a similar assumption to the generalization gap for reconstruction-based methods: the student mimics the teacher for normal data but diverges when presented with anomalous inputs. The general setup for distillation-based methods is illustrated in Figure~\ref{fig:distillation_setup}, where the student is often symmetric to the teacher, that is, sharing the same architecture.

Initial works extended the CNN-based distillation method introduced by  \citet{wang2021studentteacher} by incorporating attention mechanisms. For example, \citet{yamada2022reconstruction} introduces an attention bottleneck to constrain the student's ability to reconstruct anomalies and added an asymmetric student-teacher pair. Similarly, \citet{yan2023multiresolution} adapted DeiT~\cite{touvron2021training} strategy to distill features from pre-trained CNNs, to an AE-transformer architecture. Here, the encoder learns to regress features from the teacher, while the decoder uses these tokenized features as surrogate labels.

Transformaly~\cite{cohen2022transformaly} employs a dual-feature space approach, leveraging discrepancies between pre-trained ViT tokens and a symmetric student. Using Gaussian distributions, the score is the combination of sample likelihood in both spaces. 

For video AD, \citet{baradaran2023multi} proposes a multi-task learning framework with two distillation branches trained on complementary proxy tasks. The first branch integrates semantic segmentation and future frame prediction, while the second handles optical flow magnitude estimation. Channel and spatial attention modules refine the backbone's ability to capture motion in object parts, with an additional attention network modeling contextual objects interactions.
On a similar note, GLCF~\cite{yao2023learning} proposes a Swin-based two-branch framework to address structural and logical anomalies using four main components in a joint optimization. A local branch detects structural anomalies, while a global branch handles logical inconsistencies. Correspondence between the two branches is learned using a ViT-based local-to-global bottleneck. Two auxiliary networks perform local-to-global feature estimation and vice-versa.



Inspired by DINO~\cite{caron2021emerging}, \citet{Bozorgtabar_Mahapatra_2023} introduce a distillation strategy where differently augmented samples are passed to the transformer-based student-teacher while enforcing representation consistency. Non-salient regions are masked out using pre-trained ViT attention maps, and synthetic anomalies generated using a cut-paste-inspired strategy~\cite{li2021cutpaste} are used for training.
\citet{zhang2024feature} 
elaborated
on this approach, using the cluster-centroid of the teacher's features as the student target to constrain it on a multi-layer basis.
\citet{li2024sam} propose a two-stream ViT-based student framework, guided by a lightweight version of the foundation model SAM, MobileSAM~\cite{zhang2023faster}, targeting mobile-friendliness in industrial settings. To avoid using SAM during inference, each student distills specific capabilities of SAM. Using additional pseudo-anomalies, one learns discriminative features for normal and anomalous regions, while the second is trained on denoising reconstruction as in \citet{zhang2023destseg}. Both students share SAM mask decoder, with their features combined to generate anomaly maps.


SQUID~\cite{xiang2023squid} combines in-painting-based feature-level distillation with a memory dictionary akin to \citet{gong2019memorizing}. Masked neighborhood patches are in-painted using Transformer-based architecture, and the reconstructed features are augmented by similar features from the memory dictionary. An anomaly score is then derived using a discriminator evaluating the student’s reconstructions.


\subsection{Distribution Map and One-Class Classification}\label{subsec:distribution_map}
Distribution map methods statistically model normal behavior, mapping normal features to a desired distribution and identifying anomalies as deviations. Similarly, one-class classification (OCC) methods define normality within a distribution or hypersphere, using proximity to this structure for anomaly scoring. However, OCC methods often rely on high-quality synthetic anomalies to define the structure boundaries, with performance degrading when 
the synthesized data is insufficient.

\begin{figure}[!ht]
    \centering
    \includegraphics[width=0.99\linewidth]{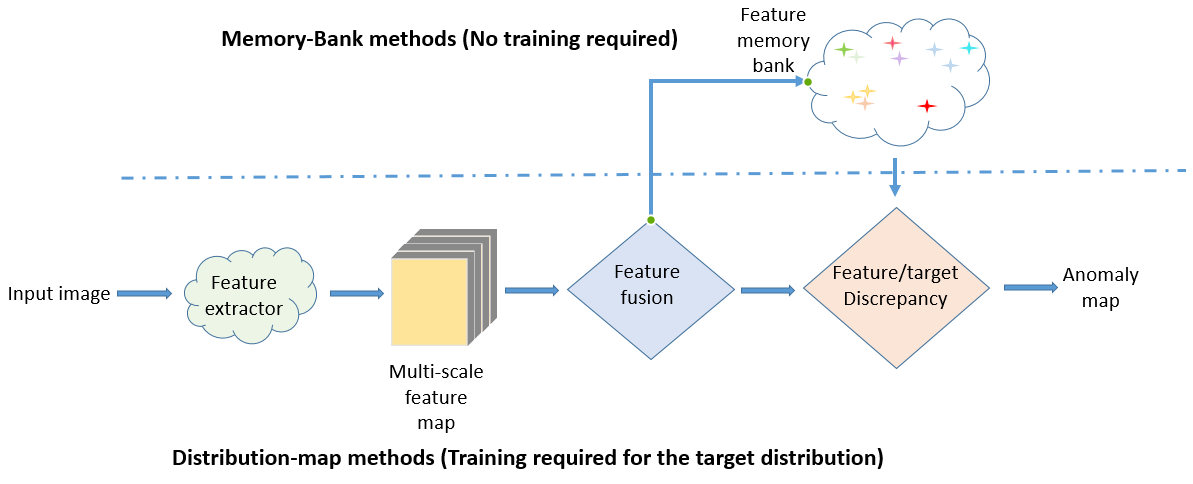}
    \caption{The basic flow of distribution-map, one-class classification, and memory-bank methods. Features are extracted using a pre-trained feature extractor, fused, and optionally stored in memory. Anomaly maps are generated based on the distance or likelihood between the test embedded vectors and the learned or stored reference vectors representing normality depending on the used approach.}%
    \label{fig:distribution-map-memory-bank-setup}
\end{figure}

OCC and distribution map methods can excel if a robust distribution map or hyper-sphere is learned. Figure~\ref{fig:distribution-map-memory-bank-setup} illustrates the core pipeline for these methods and highlights their connection to memory-bank approaches (see Section~\ref{subsec:memory-bank}).

Early methods adapted state-of-the-art algorithms like Deep Support Vector Data Description (SVDD)~\cite{cmc.2023.035246}, using ViT's class token to form a minimal-radius hypersphere enclosing normality features. Similarly, \citet{visapp23} employs pre-trained Convolutional Transformer conTNet~\cite{yan2021contnet} to extract patch-wise embeddings, modeling normality with multivariate Gaussian distributions.

For the video domain, SSMTL++~\cite{barbalau2023ssmtl++} builds on the multi-task method developed by \citet{georgescu2021anomaly}, adding proxy tasks, optical flow, and a hybrid 3D Convolutional Vision Transformer~\cite{wu2021cvt} backbone for improved performance.
\citet{chang2022video} proposed a dual-stream AE to model spatial and temporal patterns separately, combining reconstruction and clustering. The temporal stream incorporates a variance-based attention module to highlight abnormal motion patterns, while k-means clustering on both streams ensures compactness and fusion.

The advent of normalizing flows (NF), a density estimation algorithm that learns an invertible mapping from a simple, usually Gaussian distribution into a complex target distribution, has inspired several AD approaches.
CAINNFlow~\cite{yan2022cainnflow} applied NF to pre-trained features, replacing the NF MLPs with CBAM attention blocks to address spatial information loss caused by NFs flattening the feature maps~\cite{cunningham2020normalizing}.
Similarly, AttentDifferNet~\cite{silva2023attention} builds on the CNN and NF-based DifferNet~\cite{rudolph2020differnet}, improving feature distribution modeling with modular attention, incorporating squeeze-excitation and CBAM blocks at multiple encoder levels.

Other works extend NFs application in AD.
For instance, \citet{PangFusion} employ mask and label attention~\cite{9850974} into a SegFormer~\cite{xie2021segformer} AE architecture, using NF as a prediction head modulated by label attention. The model first trains on synthetic anomalies, after which its weights are frozen to fit the NF.
U-Flow~\cite{tailanian2024u} employs a CaiT Transformer~\cite{touvron2021goingdeeperimagetransformers} for multi-scale feature extraction, combining a U-shaped NF with dynamic thresholding via multiple-hypothesis testing refined scoring.
\citet{yao2024local} propose a Dual-Branch ViT for reconstructing local-global semantics, followed by a conditional NF for dual distribution modeling. The second branch utilizes reconstructed local features in the first for an aggregation-cross-attention operations with global tokens to enhance global semantics reconstruction. 
Complementarily, \citet{chiu2023self} employ synthetic anomaly data as out-of-distribution constraints for NF, employing a cascade of convolution and self-attention blocks to refine normalcy distribution learning.

Self-supervised learning (SSL) has inspired a range of innovative methods. Inspired by BarlowTwins~\cite{zbontar2021barlow}, \citet{yao2023focus} employ a two-branch architecture to focus on relational patterns. Patch-wise self-attention maps are used to generate intra- and inter-correlation maps, which serve as the SSL training targets and anomaly indicators.
Building on the popular CutPaste framework, \citet{Bozorgtabar_2022_BMVC} integrate attention maps to select pasting regions. Test-time-augmentation (TTA) is employed to align the attention maps statistics at the training and inference stages.

In the video domain, self-supervision is primarily used to generate pseudo anomalies.
 \citet{wu2022self} leverage the pseudo-anomalies to jointly optimize two complementary attention-based modules for classification. The first uses dictionary learning~\cite{mairal2009online} to filter anomaly signals, while the other isolates anomalies through selective filtering.
Subsequently, CIForAD~\cite{li2024cross} incorporates textual prompts (see section~\ref{sec:zero_few_shot} for more details), paired with a Transformer-based backbones.
The model comprises three key components, with their combination providing the final score:(1) bi-directional encoding to enhance visual-textual feature fusion, (2) semantic correlation alignment between modalities through local training and global inference, and (3) a time-attention module leveraging SSL-generated pseudo-anomalies to capture inter-frame discrepancies.
Focusing on complex and industrial scenarios, ~\citet{liu2024ampnet} propose \textit{AMP-Net}, a dual-stream appearance-motion prototype network. AMP-Net leverages multiscale spatial fusion, variance-based temporal attention, and a spatial-temporal fusion module with external memory banks to learn normalcy prototypes. By fusing learned representations with cross-attention and enhancing them through adversarial training, the model balances discriminative normality learning with limited anomaly generalization.

However, these methods often assume that the distribution of normal features is stationary and disentangled from environmental or scene-specific variations, which is rarely the case in real-world multi-scene surveillance settings.
To explicitly address this issue, ~\citet{liu2025crcl} propose CRCL (Causal Representation Consistency Learning), which leverages a structural causal model to disentangle causal normality from scene bias, by combining scene-debiasing and motion-aware causal normality learning.

\subsection{Memory-Bank and Feature Matching}\label{subsec:memory-bank}

Memory bank-based methods store pre-trained representative features to model normality, and identify anomalies as significant deviations from this reference. Their effectiveness hinges on feature quality and memory size, but a well-constructed memory bank can achieve state-of-the-art performance.
Figure~\ref{fig:distribution-map-memory-bank-setup} illustrates the core pipeline for these methods, highlighting their connection to other feature-based methods.

Early attention-based memory bank methods drew inspiration from CNN-based approaches. SA-PatchCore~\cite{10007829} extends PatchCore~\cite{Roth_2022_CVPR} by incorporating a modified MHSA module for mid-level feature extraction, enhancing the memory bank with co-occurrence and logical anomaly awareness.
Inspired by the same method, SelFormaly~\cite{lee2023selformaly} utilizes ViT’s attention maps to filter background patches, optimizing memory efficiency.
MemAE~\cite{gong2019memorizing} introduces a sparse normality-prototype memory bank, using encoder’s representations as queries to retrieve similar prototypes for reconstruction, effectively filtering anomalous signals.

\citet{liu2023component} propose a lightweight plug-in method to mitigate limitations with logical anomalies detection. The framework uses DINO pre-trained features to build a memory bank through coreset sub-sampling and KMeans clustering. This bank is used to generate the query image component-level segmentation map, which are utilized to extract metrological features such as size, quantity and color. Logical anomalies are detected through a component-based binary classification on these features.

Memory banks have also been adapted for multimodal scenarios, integrating features from diverse domains.
Multi-3D-Memory~\cite{wang2023multimodal} proposes a three step approach to fuse 3D point clouds and 2D RGB images: (1) domain-specific features are extracted using DINO~\cite{caron2021emerging} for 2D and PointMAE~\cite{pang2022masked} for 3D; (2) the domains are aligned and fused via interpolation and projection on the 3D features, followed by an unsupervised contrastive learning step; and (3) three memory banks are created---one per domain and one for fused features—used by Support Vector Machines for anomaly scoring.
\citet{wang2023investigating} refines this approach for resource-constrained devices by incorporating post-training model quantization and pruning, improving computational efficiency.

\subsection{Discussion}
\label{discussion_feature}






Capitalizing on the representational power of large-scale pre-trained models to extract high-quality features, feature-based methods have emerged as a prominent approach for anomaly detection (AD). 
Utilizing the premise that anomalies manifest as deviations in the feature space to the tightly clustered normal samples. Feature methods enable greater customization and efficiency with fewer parameters than reconstruction-based approaches, with the integration of Transformers further enhancing their adaptability and performance.

The primary advantage of feature-based methods lies in their ability to harness pre-trained representations, which often capture rich semantic information without requiring extensive training on domain-specific datasets. This makes them highly data-efficient and suitable for low-data regimes. Additionally, feature-based methods are more generalizable compared to reconstruction-based approaches thanks to their reliance on high-level semantic representations. Memory bank methods, in particular, have demonstrated strong performance in both image and video anomaly detection by offering a compact and discriminative feature space. 
On the other hand,
distillation methods benefit from diverse optimization targets, aligning closely with self-supervised Transformer-based representation learning strategies.

Despite their strengths, feature methods are heavily dependent on the quality of the pre-trained feature extractor, which may not always align with the AD task. Furthermore, these methods often require careful design of distance metrics or statistical models to effectively separate normal and anomalous samples. In addition to their higher inference time compared to reconstruction methods, the lack of explicit reconstruction makes it more difficult to localize anomalies. Additionally, memory bank methods are prone to scalability issues, as the size of the memory bank grows with the diversity of normal data.
Balancing generalization and specialization is another challenge, as models must perform well across diverse datasets while maintaining high precision on specific anomaly types. Interpretability remains limited, with few methods offering insight into the decision-making process of feature-based anomaly detection. Moreover, uncertainty estimation is often overlooked, 
although it
could improve confidence calibration and enhance decision reliability.

Looking forward, combining pre-trained networks with fine-tuned lightweight models could improve generalization while maintaining computational efficiency. 
The particular synergy between distillation methods and self-supervised Transformer-based representation learning algorithms may open pathways for more robust teacher networks and AD algorithms. Local-global feature interactions may be further enhanced leveraging specifically tailored attention modules, particularly for logical anomalies that demand a nuanced understanding of feature relationships.
Additionally, the development of multi-modal feature representations, incorporating visual, temporal, and semantic cues, could lead to more robust AD systems. 
As the field progresses, addressing these challenges will be essential to fully exploit the potential of feature-based methods.

\section{Foundation Models \& Zero- and Few-Shot Anomaly Detection}
\label{sec:zero_few_shot}
Zero- and Few-Shot Anomaly Detection (ZSAD/FSAD) enables anomaly detection with minimal or no training data, addressing challenges in rare anomaly scenarios, privacy constraints, and rapid adaptation to new products.

Large pre-trained Vision-Language Models (VLMs)~\cite{radford2021learning,li2022blip,jia2021scaling,alayrac2022flamingo}, Large Language Models (LLMs)\cite{ouyang2022training, touvron2023llama,raffel2020exploring,openai2024gpt4} and large multimodel models (LMMs)~\cite{yang2023dawn,girdhar2023imagebind,awadalla2023openflamingo} demonstrate strong zero-shot capabilities, inherently supporting multi-class AD (see Section~\ref{Multi-class}) without category-specific fine-tuning. They also enhance interpretability by providing textual explanations.
However, challenges remain despite this potential, particularly in detecting subtle and domain-specific abnormalities. This is highlighted by the LLMs benchmark for industrial anomaly detection provided by~\cite{jiang2024mmad}.

Despite excelling in associating text prompts with global image representations, foundation models tend to prioritize class semantics over subtle anomaly detection. Their reliance on general datasets further limits their ability to capture domain-specific details. While addressing these challenges is crucial to benefit from their capability, direct fine-tuning on AD datasets is prone to overfitting, necessitating alternative strategies.

To mitigate these challenges, ZSAD methods commonly employ prompt-embedding  techniques~\cite{zang2022unified,jia2022visual, ju2022prompting,Zhou_2022,shen2022multitask} to adapt textual prompts for AD tasks, aligning with the text-dominant predictions of VLMs.
Hence, prompt quality plays a critical role in performance.
Another approach is to incorporate a small number of trainable parameters, enabling adaptation with auxiliary AD data while avoiding overfitting.

\begin{figure}[!ht]
    \centering
    \includegraphics[width=0.8\linewidth]{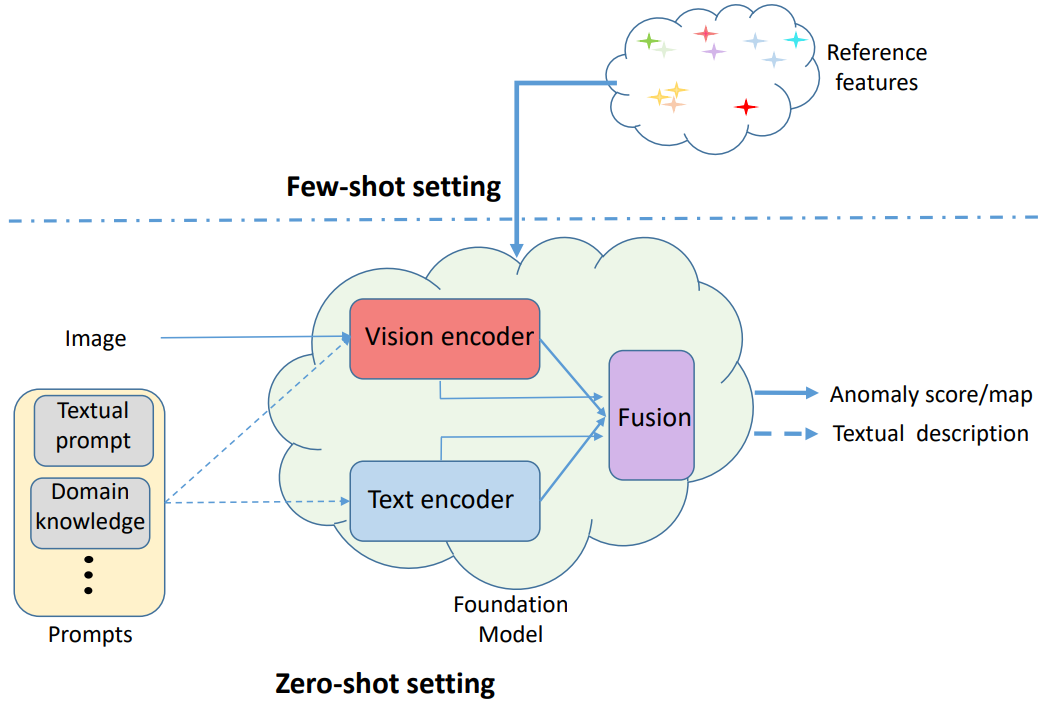}
    \caption{The basic flow of foundation models-based zero- and few-shot methods. The image and prompts (which can be learned) are processed by the model, with anomaly scoring determined through modality alignment.  A small number of learnable parameters may be integrated for task-specific adaptation using auxiliary AD datasets. In few-shot settings, a memory bank can be added to aid in prediction. }%
    \label{fig:foundation_fig}

\end{figure}

These ZSAD methods typically compute anomaly scores based on the similarity between visual tokens and textual prompts.
Training-free adaptations, such as memory banks for test-time comparisons, can be integrated in ZSAD methods to further enhance performance methods in a few-shot settings. Figure~\ref{fig:foundation_fig} illustrates an overview of these methods.

This section categorizes ZS/FSAD solutions into three main groups:
(1)~Methods leveraging CLIP and SAM due to their open-source nature and widespread use.
(2)~Other foundation model-based approaches, extending beyond CLIP and SAM.
(3)~Transformer-based methods bypassing foundation models.

It may be worth noting that, while several recent works \citep{wei2022look,wei2022msaf,joo2023cliptsaclipassistedtemporalselfattention,wu2023vadclipadaptingvisionlanguagemodels} have integrated multimodal cues such as audio and text for video anomaly detection, most operate under supervised or weakly-supervised settings. These approaches are not explored in this review as we focus on unsupervised VAD. In contrast, most unsupervised multimodal methods remain limited to text-vision alignment, as enabled by foundation models like CLIP or dual-branch VLMs.

\subsection{CLIP \& SAM Based}\label{subsec:CLIP_based} 
The open-source availability of foundation models like CLIP and SAM has driven their adoption for ZSAD, particularly CLIP. SAM primarily enhances segmentation by providing general masks, while CLIP's prompt-based adaptations leverage its visual-text alignment for anomaly detection.
However, CLIP’s reliance on global embeddings can weaken fine-grained localization. To address this, modifications such as value-value (V-V) attention~\cite{li2023clip} focus solely on value vectors, reducing the averaging effect of query-key interactions and improving anomaly sensitivity.

WinCLIP~\cite{jeong2023winclip} enhances CLIP's domain-specific knowledge with object-state prompts, pairing binary state descriptors (e.g., normal/anomalous) with unified text templates. It refines localization by computing anomaly scores over overlapping, multi-scale windows, mitigating global embedding bias.
Expanding on this, \citet{baugh2023zeroshot} integrates SAM~\cite{kirillov2023segment} and CLIPSeg~\cite{lüddecke2022image} into WinCLIP framework, where SAM extracts object masks, and CLIPSeg refines pixel-level predictions for better anomaly localization.
SAA++~\cite{cao20232nd} further improves performance by cascading foundation models. GroundingDINO detects anomaly regions, refined by SAM, while multimodal prompts incorporating domain-specific knowledge and visual context improve calibration and localization.

Another challenge in CLIP-based ZSAD is the potential overlap between normal and anomalous prompts embeddings. To address this, some methods emphasize separating their representations.
 CLIP-AD~\cite{chen2024clip} selects representative text prompts from learned normal and abnormal distributions. Its vision branch employs V-V attention and feature surgery~\cite{li2023clip}, with an optional fine-tuning step of CLIP’s linear layers to optimize text-image alignment.
ALFA~\cite{zhu2024llms} dynamically separates prompt embeddings by leveraging GPT-3.5~\cite{openai2022chatgpt} at runtime to generate context-aware prompts. Starting with an expert-informed prompt, ALFA adapts it per sample, incorporating V-V attention to refine local-global semantic alignment.

Several methods integrate trainable parameters into CLIP to enhance text-image alignment, localization, and prompt quality for anomaly detection.
\citet{tamura2023random} extends WinCLIP by adding a learnable feedforward neural network (FNN) atop CLIP’s encoders to enhance modality alignment. Instead of predefined object categories, it generates prompts using CLIP’s text encoder, augmenting them with random words to increase diversity for the FNN training.
APRIL-GAN~\cite{chen2023aprilgan} uses additional linear layers to project
 visual and textual features into a shared embedding space. The used prompts are derived from CLIP’s training templates and augmented with object-state information.
Similarly, AnoVL~\cite{deng2023anovl} employs a trainable layer for Test-Time Augmentation, improving text-patch alignment for better localization. It further refines CLIP’s visual encoder with V-V attention and domain-aware contrastive prompting.
AnomalyCLIP~\cite{zhou2023anomalyclip} shifts focus from object semantics to anomaly states, using learnable, object-agnostic prompts to represent normality and abnormality.  These undergo global-local context optimization to align with multi-scale visual features, while V-V attention further refines textual and visual spaces.

PromptAD~\cite{Li_2024_WACV} introduces a dual-branch CLIP-based architecture, where each branch models either normal or anomalous states guided by textual prompts. During training, inter-branch dissimilarity is used as an attention map to guide the opposing branch, with the final anomaly map aggregating both branches predictions.

Since prompt quality is crucial for CLIP’s performance in anomaly detection, several methods focus on their refinement.
AdaCLIP~\cite{cao2025adaclip} integrates learnable prompting layers~\cite{khattak2023maple} to generate hybrid prompts. these prompt include static-tokens which are shared across images and dynamic- image specific-tokens at test time, ensuring both global and instance-specific prompting. A learnable projection layer aligns these textual prompts with WinCLIP’s templates for improved text-image coherence.

Similarly, FiLo~\cite{gu2024filo} leverages LLMs to generate adaptive textual prompts that identify potential anomaly types. GroundingDINO~\cite{liu2023grounding} then removes background interference by incorporating positional information into text prompts for alignment.Additionally, FiLo uses V-V attention and variable-size convolutions for multi-scale text-image alignment, with an adapter to optimize global feature alignment.
ClipSAM~\cite{li2025clipsam} fuses CLIP and SAM into a two-stage pipeline: CLIP performs initial segmentation, prompting SAM for refined localization. CLIP’s image-text alignment is refined using two learnable attention-based pathways, emphasizing both local and global alignment, while textual prompts are adapted from WinCLIP’s framework.
Extending CLIP to videos, \citet{kim2023unsupervised} use predefined prompts to describe frames via GPT-4~\cite{openai2023gpt4}. These descriptions, enriched with domain knowledge, serve as CLIP prompts. An additional layer atop CLIP is fine-tuned contrastively to enhance alignment.

For FSAD, additional techniques enhance performance by incorporating auxiliary samples or structural constraints.
InCTRL~\cite{zhu2024toward} incorporates trainable layers into CLIP’s vision encoder to learn multi-level tokens and global image residuals. Using WinCLIP’s prompt templates for modality alignment, the final score is refined with the residual computed against the auxiliary few-shots.  

\citet{li2024promptad} proposes Semantic Concatenation (SC) and Explicit Anomaly Margin (EAM), combined with V-V attention, to improve localization. SC transforms normal prompts into anomaly-aware prompts by appending anomaly-specific suffixes, enabling contrastive learning. EAM imposes a margin constraint to explicitly separate normal and anomalous features in the absence of labeled anomaly samples.
SAM-LAD~\cite{peng2024sam} addresses logical anomalies by building a feature bank using DINOv2~\cite{oquab2023dinov2} and leveraging SAM for query object segmentation. Query objects undergo k-nearest neighbor matching against the memory bank, with unmatched objects signaling logical anomalies. Structural anomalies are further handled using Multivariate Gaussian Distributions (MVGs) to model object-level feature distribution discrepancy.

\subsection{Vision-Language Models Beyond CLIP \& SAM for ZSAD}\label{subsec:other_foundation}

Beyond CLIP, various VLMs have been explored for ZSAD by aligning vision and text modality representations. These approaches often combine foundational vision models with LLMs, providing explainability through textual description of anomaly regions alongside reasoning for model predictions.

AnomalyGPT~\cite{gu2023anomalygpt} enables interactive anomaly detection without requiring manual thresholding. It uses ImageBind-Huge~\cite{girdhar2023imagebind} as the vision encoder and Vicuna-7B~\cite{zheng2023judging} as the LLM. Fine-tuned on synthetic anomalous visual-textual data, AnomalyGPT aligns vision and language modalities through a feature-matching decoder.
Similarly, Myriad~\cite{li2023myriad} combines MiniGPT-4~\cite{zhu2023minigpt} with a pre-trained Q-Former~\cite{li2023blip2}, a domain adapter, and a visual instructor. These components refine anomaly maps before fusing them with LLM-generated outputs, improving both detection performance and interpretability.

\citet{cao2023towards} evaluate GPT-4V~\cite{yang2023dawn} across multiple domains using structured prompts. These prompts incorporate task details, class specifications, normalcy standards, and reference images with SAM-generated image masks assisting in anomaly localization. While GPT-4V was shown to be effective qualitatively, this approach lacks comprehensive quantitative evaluation.
Conversely, \citet{zhang2023exploringVQA} provide quantitative results, revealing that despite its promise, GPT-4V trails behind CLIP-based methods in pixel-level localization. This study uses a Visual Question Answering prompting framework, segmenting images into superpixels and prompting GPT-4V to localize anomalies within these divisions.

Adaptations to the video domain typically involve frame-level captioning, with a focus on temporal consistency.
TEVAD~\cite{chen2023tevad} uses dense video captions generated by SwinBERT~\cite{lin2022swinbert}. These captions are embedded within a textual branch and fused with ResNet-based visual features, forming a dual-branch model that enables transparent and explainable detection.

Subsequent works on video data shifted to generating anomaly scores using LLMs.
AnomalyRuler~\cite{yang2025follow} introduces a rule-based reasoning framework for one-class, few-shot detection, consisting of two stages: (1) Rule Generation; normal video samples are analyzed using CogVLM-17B~\cite{wang2023cogvlm}, generating frame-level textual descriptions to prompt GPT-4 to generate a rule set for common pattern. (2) Anomaly Detection; Mistral-7B~\cite{jiang2023mistral} applies the learned rules on its test-samples textual descriptions, providing detection with detailed reasoning.
LAVAD~\cite{zanella2024harnessing} proposes a training-free approach, using BLIP-2~\cite{li2023blip2} to provide captions, which are then denoised through cross-modal alignment with visual embeddings. These captions are temporally grouped and used as prompts for LLaMA~\cite{touvron2023llama} to provide a textual anomaly score. The final score consists of the alignment between the LLM scores and the aggregated frame embeddings from ImageBind video encoder, enhancing temporal and spatial consistency.

\subsection{Non-Foundation Model Approaches for Zero- and Few-Shot Anomaly Detection}\label{subsec:FS_without_foundation}

While foundation models dominate ZSAD and FSAD research, some methods address these tasks without relying on them. Instead, they leverage strong prior representations and innovative modeling strategies. Given the significant gap in representation power compared to foundation models, these efforts primarily focus on FSAD.

\citet{Takimoto2022} proposes a metric-based approach using an attention-augmented Convolutional Siamese network. To counteract the scarcity of anomalous data, the method incorporates weighted contrastive learning, which emphasizes task-specific features using attention modules inspired by \citet{Fukui_2019_CVPR}. This design ensures that the network highlights critical anomaly-relevant regions.

Other FS methods addressed the multi-class task.
Metaformer~\cite{9711445} employs a transformer-based instance-aware AE optimized via unsupervised meta-learning. This approach involves training on few-shot reconstruction tasks using auxiliary datasets, followed by fine-tuning to the target domain.
\citet{huang2022registration} use feature registration as a category-agnostic proxy task. Registration transforms different images into a shared coordinate system to enable better comparison, maximizing intra-category feature similarity.
 A spatial transformer-based Siamese network generates these alignments, constructing a normalcy memory bank. Anomalies are detected as deviations from learned MVG distributions over registered features.

In the video domain, FewVAD~\cite{fakhry2024enhancing} incorporates an attention-based selection module to filter redundant frames using mutual information. A Spatial-Temporal Cross Transformer is then applied, where: spatial attention focuses on relevant regions within frames, temporal attention identifies key time intervals. A cross attention operation references these features 
against few-shot video prototypes, with anomalies detected based on dissimilarities.

Though less common, a few methods tackle ZSAD without foundation models.
MUSC~\cite{li2024musc} builds on the observation that normal samples exhibit high intra-similarity compared to anomalies. Therefore, they propose to contrast pre-trained ViT patch embedding on the unlabeled test set among themselves. Various scale anomalies are captured by aggregating multi-level image patches with varying neighborhood sizes, with the scores assigned based on immediate image-neighborhood similarity. To reduce sensitivity to noise, scores are reweighted using a constrained graph-based similarity measure derived from class token embeddings to access image-level semantics.
MAEDAY~\cite{schwartz2022maeday} exploits the strong representation capabilities of pre-trained Masked Autoencoders (MAEs, see Section~\ref{subsec:MAE}). By averaging input reconstruction multiple times with random masks, MAEDAY detects anomalies without requiring labeled data. The model can be fine-tuned with a reconstruction objective in few-shot settings. 
The authors further highlight that the knowledge learned through reconstruction is orthogonal to feature-based methods. Hence, an enhanced performance can be obtained when combining both approaches, an insight utilized by many methods throughout this survey.

\subsection{Discussion}
\label{discussion_ZS_FS}

Foundation models and zero- and few-shot anomaly detection (ZSAD/FSAD) have emerged as transformative paradigms, addressing the long-standing challenge of detecting anomalies with minimal or no labeled training data. These methods leverage large pre-trained vision-language models (VLMs), large language models (LLMs), and multimodal models that inherently support multi-class anomaly detection without the need for category-specific fine-tuning. By aligning visual representations with textual prompts, foundation models offer a unique capability to generalize across a broad range of tasks while enhancing interpretability through textual explanations. This adaptability makes foundation models particularly attractive for applications requiring rapid adaptation and transparency. Furthermore, lightweight prompt engineering, memory banks and fine-tuning strategies have extended their applicability to few-shot settings, enabling higher performance with limited data.
Non-foundation model approaches, on the other hand, offer competitive alternatives by leveraging task-specific architectures and metric-based learning, demonstrating strong performance in low-data regimes and constrained environments.

Despite their strengths, foundation models face notable challenges. Their reliance on large-scale general-purpose pre-training data often limits their ability to detect subtle or domain-specific anomalies. Additionally, their high computational demands pose practical barriers to deployment in resource-constrained environments. Overfitting remains a critical concern when fine-tuning these models on small datasets. Furthermore, the interpretability of their anomaly scores is often limited to coarse textual explanations, lacking the fine-grained localization required for many applications.
Open challenges in foundation model-based AD include scalability, computational efficiency, and uncertainty estimation. Developing methods that adapt foundation models to domain-specific tasks without excessive fine-tuning remains an active area of research. Balancing the trade-off between broad generalization and high precision on specific anomaly types is another pressing issue. Additionally, improving the interpretability of foundation model-based methods through explainable AI techniques is crucial for fostering trust in their decision-making process.

The future of ZSAD and FSAD is likely to be driven by hybrid approaches that combine the broad representational power of foundation models with the domain-specific efficiency of other deep learning algorithms. Few-shot prompting strategies, self-supervised learning, and uncertainty-aware inference mechanisms will play a key role in enhancing the reliability of these systems. Furthermore, the development of lightweight foundation model architectures and knowledge distillation techniques could enable deployment on edge devices. As these methods mature, they have the potential to revolutionize AD across a wide range of applications, from industrial inspection to medical diagnostics, by providing scalable, interpretable, and data-efficient solutions.

When using foundation models, particularly those pretrained on large-scale web datasets such as LAION-400M~\cite{birhane2021multimodal}, an important aspect to consider is that they often inherit significant biases reflecting demographic, cultural, and scene imbalances~\cite{zhao2021understanding, bandy2021addressing}. In visual anomaly detection (VAD), such biases can lead to high false positive rates in critical domains such as medical imaging or industrial inspection, where normal patterns may deviate from web-scale distributions~\cite{oakden2020exploring, zhang2023industrial}. These risks are especially pronounced in unsupervised settings, where no labeled anomaly data is available to recalibrate model behavior~\cite{ruff2021unifying, scholkopf2021toward}. 
For example, vision-language models like CLIP \cite{radford2021learning} have demonstrated impressive zero-shot transfer, yet studies have shown their performance degrades significantly on out-of-distribution or fine-grained tasks \cite{devries2022clip, shen2023benchmark}. Addressing these concerns requires targeted bias mitigation strategies. Prompt engineering \cite{zhou2022learning}, domain-adaptive fine-tuning \cite{kumar2022domain}, and bias-aware evaluation \cite{buolamwini2018gender, zhang2023fairness} are increasingly crucial as foundation models become integral to the VAD pipeline.

\section{Conclusion}

Transformers and foundation models have transformed visual anomaly detection (VAD), overcoming key limitations of traditional convolutional architectures. Their ability to capture global dependencies, adapt through prompt engineering, and excel in unsupervised and self-supervised learning paradigms has driven a surge of diverse innovative solutions. 
Reconstruction- and feature-based methods benefit from Transformers representational power and flexibility, while foundation models extend this with zero- and few-shot capabilities, and enhanced interpretability.
Foundation models, particularly those based on vision-language alignment, enable zero- and few-shot capabilities while enhancing interpretability through natural language prompts.

Despite these advancements, challenges persist, particularly in detecting subtle or domain-specific anomalies, and improving data efficiency. Additionally, the interpretability of these models often remains limited, particularly in explaining fine-grained anomaly localization decisions, while uncertainty estimation is mostly overlooked.

An interesting avenue for future research is to prioritize optimizing models for low-data environments and exploring hybrid approaches that combine the strengths of Transformers, foundation models, and CNNs. Bridging these gaps while producing uncertainty-aware algorithms could unlock more scalable and effective VAD systems, expanding their impact across diverse real-world domains.



\clearpage


\bibliography{iclr2025_conference}
\bibliographystyle{iclr2025_conference}

\end{document}